%% file: main.tex
\documentclass[letterpaper]{article}
\usepackage[preprint]{aaai2027}
\usepackage[hyphens]{url}
\usepackage{graphicx}
\usepackage{natbib}
\usepackage{booktabs}
\usepackage{amsmath,amssymb}
\title{EviDent-CBCT: Evidence-Bottlenecked Report Generation from Dental CBCT under Non-Exhaustive Report Supervision}

\author{Ruiyang Hao, Zhi Qin Tan, Yulan He, Owen Addison, Yunpeng Li\thanks{Corresponding author. This work was supported by King's College London via the Faculty of Dentistry, Oral and Craniofacial Sciences Newland Pedley Studentship.}}

\affiliations{
King’s College London\\
\{ruiyang.hao, zhi\_qin.tan, yulan.he, owen.addison, yunpeng.li\}@kcl.ac.uk}

\begin{document}
\maketitle

\begin{abstract}
Dento-maxillofacial cone-beam CT (CBCT) reports may contain dozens of tooth-specific, anatomical, and spatial findings from a single 3D scan. Learning to generate such reports from limited clinical data is challenging because routine reports may not exhaustively document image findings, and a non-mention may reflect either absence or non-reporting. We present EviDent-CBCT, an evidence-bottlenecked framework designed for this incomplete supervision. An anatomy-aware network maps each CBCT scan to a discrete record of tooth-level, global, and tooth-IAC evidence. A dental-logic consistency projection reconciles incompatible evidence before a deterministic renderer and an image-blind local language model generate the report using only this record. For tooth-level evidence, reliability-aware training uses eligible non-mentions as reduced-weight negatives, while unreported global and tooth-IAC labels remain unknown. A metal-sensitive input channel preserves intensity cues from dental materials.
Across three validation runs, EviDent-CBCT achieves \(0.666\pm0.006\) merged evidence set-F1 and \(0.402\pm0.003\) RadFact-Lite-Dental logical-F1, versus \(0.371\pm0.018\) for the strongest controlled direct baseline. In the ODIN 2026 challenge, it ranked second in automated evaluation and third in blinded clinical Arena comparison on the hidden test set. These results support the discrete evidence record as an effective and auditable interface for CBCT report generation.
\end{abstract}

\input{sections/1_intro}

\input{sections/2_related_work}
\input{sections/3_problem}
\input{sections/4_method}
\input{sections/5_experiments}
\input{sections/6_conclusion}
\small{
\bibliography{references}
}

\input{sections/7_supplementary}

\end{document}

%% file: sections/1_intro.tex
\begin{figure*}[t]
    \centering
    \includegraphics[width=\textwidth]{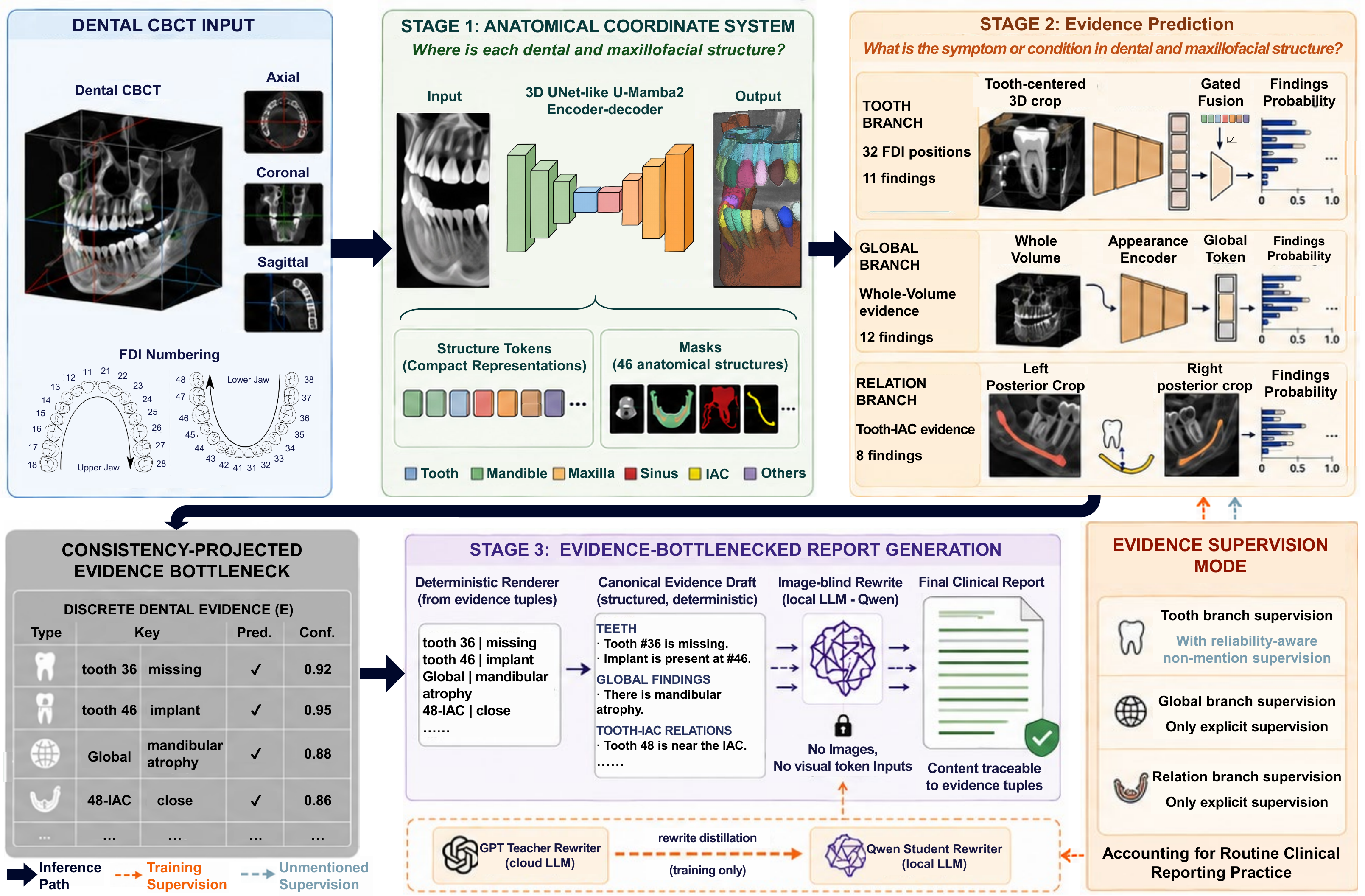}
    \caption{\textbf{Overview of EviDent-CBCT.} During training, routine clinical reports are converted into source-linked reference evidence that preserves verbatim supporting spans; explicit labels supervise all branches; eligible tooth-level non-mentions provide reduced-weight negative supervision. At inference, Stage 1 extracts anatomical masks and structure tokens from the CBCT volume. Stage 2 combines tooth-centered, whole-volume, and posterior-mandibular views to predict tooth, global, and tooth-IAC findings, which are consolidated into a discrete evidence record. A dental-logic consistency projection reconciles incompatible evidence before Stage 3 deterministically renders a canonical draft, which serves as the sole case-specific input to an image-blind local Qwen rewriter. The rewriter is trained through GPT-based content-preserving distillation on drafts constructed from consistency-projected predicted evidence. Solid arrows indicate inference; orange arrows signify training-only supervision; and light-blue arrows denote reduced-weight supervision from eligible tooth-level non-mentions.}
    \label{fig:pipeline}
\end{figure*}

\section{Introduction}

Cone-beam computed tomography (CBCT) provides three-dimensional views of dento-maxillofacial structures and is widely used for dental diagnosis and treatment planning~\cite{kaasalainen2021dental,reddy2025integration}. A single scan may reveal findings involving individual teeth, the jaws, sinuses, scan coverage, and spatial relations to the inferior alveolar canal (IAC), often requiring assignment to a specific anatomical region or FDI tooth position. CBCT reporting is therefore more than image captioning: a system must recover case-specific findings, preserve their anatomical identity and polarity, and express them coherently.

Routine reports, however, record what was documented in clinical practice rather than exhaustively annotating the image. A non-mention may indicate true absence, report-specific omission, habitual non-reporting, or oversight, and these mechanisms cannot be distinguished from the report alone~\cite{pandharipande2016rethinking,hartung2020create}. Nevertheless, aggregate non-mention patterns may provide useful negative signal without establishing absence in an individual case. We refer to this setting as non-exhaustive routine report supervision; its scope and interpretation are detailed in supplementary Sect.~A.

This setting is particularly challenging for direct image-to-text generation. Recent radiology systems connect visual encoders to large language models~\cite{wang2023r2gengpt,li2024kargen,liu2024bootstrapping,xiao2025radiology}, but with limited CBCT-report pairs, a direct generator must jointly learn three-dimensional anatomy, visual grounding, report-specific content selection, and language generation from incomplete targets. These functions remain entangled, obscuring whether candidate-reference discrepancies arise from visual prediction, incomplete documentation, or realization. Canonicalization reduces linguistic variation~\cite{vosshenrich2023investigating,bannur2023learning} but not sparse visual supervision; auxiliary evidence losses do not ensure that generation depends on predicted evidence; and fixed templates provide control at the cost of expressive flexibility~\cite{delbrouck2025automated,li2025s}. These limitations motivate an explicit content interface that separates evidence prediction from language realization and prevents the decoder from bypassing predicted findings.

We propose EviDent-CBCT, a three-stage framework built around a discrete evidence bottleneck (Figure~\ref{fig:pipeline}). Stage~1 segments individual teeth and key anatomical structures and extracts structure-specific features. Stage~2 uses anatomy-aligned branches to predict tooth-level, global, and tooth-IAC evidence; eligible tooth-level non-mentions provide reduced-weight negative supervision, unreported global and tooth-IAC labels remain unknown, and a metal-sensitive channel preserves dental-material intensity cues. A deterministic dental-logic projection then reconciles incompatible tuples using anatomical semantics and confidence. Stage~3 renders the corrected record into a deterministic draft and supplies only this draft to an image-blind language model. Thus, unlike direct generators conditioned on continuous visual representations, EviDent-CBCT makes the consistency-projected record the sole source of case-specific information, preventing bypass while keeping content selection and rewrite-induced deviations auditable. Across three validation runs, EviDent-CBCT achieves \(0.402\pm0.003\) RadFact-Lite-Dental logical-F1, versus \(0.371\pm0.018\) for the strongest controlled direct interface. It ranked second in the automated evaluation and third in blinded clinical comparison in the ODIN 2026 challenge. Our main contributions are:

\begin{itemize}
    \item We introduce an evidence-bottlenecked framework in which a discrete evidence record is the only case-specific input to an image-blind report generator, supporting transparent dental-logic correction and an auditable record of selected case-specific content.
    
    \item We design anatomy-aligned evidence learning under non-exhaustive report supervision, using class-dependent reliability weights for tooth-level non-mentions and a metal-sensitive channel for dental intensity cues.
    
    \item We provide controlled direct-generation comparisons across five visual interfaces, two report targets, and general-purpose and medically aligned language decoders, together with three-run tooth-branch ablations and a reference-derived analysis of evidence coverage, prediction, and realization.
\end{itemize}

%% file: sections/2_related_work.tex
\section{Related Work}

\paragraph{Medical report generation.}
Medical report generation has evolved from recurrent and joint disease-report models \citep{jing2018automatic,wang2018tienet} to memory-based, retrieval-based, and region-guided architectures \citep{chen2020generating,chen2021cmn,endo2021cxrepair,tanida2023rgrg}. Recent systems increasingly connect visual encoders to large language models through frozen-LLM adaptation, diagnosis prompts, knowledge augmentation, and flexible token fusion \citep{wang2023r2gengpt,liu2024bootstrapping,jin2024promptmrg,li2024kargen,wang2025llmrg4,goswami2025medivlm,shi2026u}. Most prior work targets chest radiographs, while CT2Rep extends direct generation to volumetric CT \citep{hamamci2024ct2rep}. Despite these advances, report decoders generally remain conditioned on continuous visual features, retrieved text, or soft diagnostic guidance.

\paragraph{Factuality and evidence interfaces.}
Factual correctness has consequently become a central concern. Prior work optimizes clinical coherence and entity, relation, or semantic-level agreement \citep{liu2019clinical,miura2021factual,delbrouck2022semantic}; RadGraph supports structured clinical extraction \citep{jain2021radgraph}, and RadFact evaluates reports through bidirectional proposition entailment \citep{bannur2024maira2}. Grounding, preference optimization, and reasoning further improve visual attribution and factuality \citep{bannur2024maira2,xiao2025radiology,jing2026reason}, but do not require generation to proceed through an explicit record of predicted findings. Structured approaches such as KERP, region-guided generation, and concept bottlenecks instead separate content prediction from verbalization \citep{li2019kerp,koh2020cbm,tanida2023rgrg}. EviDent-CBCT makes this separation a hard inference-time constraint: an auditable evidence record is the sole case-specific input to an image-blind realizer.

\paragraph{Dental CBCT Analysis.}
Dental CBCT learning has primarily focused on anatomical segmentation. nnU-Net provides a strong general framework \citep{isensee2021nnunet}; connectivity-aware methods address the thin inferior alveolar canal \citep{liu2023toothfairy,bolelli2024segmenting}; and recent work expands multi-structure segmentation through large dental benchmarks and state-space models \citep{bolelli2026multi,tan2025umamba2}. Moving toward clinical interpretation, Lumetti et al.\ formulate dental CBCT reporting as ontology-grounded prediction of finding presence, anatomical allocation, and clinical qualifiers from clinician-validated structured targets \citep{lumetti2026ontology}. Their system treats structured prediction as the final output and does not address narrative realization; its task-specific schema and structured metrics are therefore not directly comparable to free-text report generation. EviDent-CBCT instead learns tooth-level, global, and tooth-IAC evidence from selective, non-exhaustive routine reports and makes the predicted discrete record the required case-specific input to an image-blind narrative realizer.

%% file: sections/3_problem.tex
\section{Problem Formulation}
Let $x$ denote a 3D CBCT volume and $y$ its observed clinical report. Direct generation estimates $p(y\mid x)$. We instead derive a finite evidence record
\begin{equation}
E^y=(E^y_{\text{tooth}},E^y_{\text{global}},E^y_{\text{relation}})
\end{equation}
from the report. $E^y_{\text{tooth}}$ contains findings indexed by FDI tooth position, $E^y_{\text{global}}$ contains volume-level findings, and $E^y_{\text{relation}}$ contains tooth-IAC and canal-course findings. The superscript emphasizes that this is evidence observed through $y$, not exhaustive image truth. Reporting is factorized as
\begin{equation}
\begin{aligned}
\widehat{E}&=q_{\theta}(x), &
\widetilde{E}&=\Pi_{\mathrm{dent}}(\widehat{E}),\\
y_0&=R(\widetilde{E}), & \widehat{y}&=G_{\phi}(y_0),
\end{aligned}
\end{equation}
where $q_{\theta}$ predicts evidence, $\Pi_{\mathrm{dent}}$ is a deterministic dental-logic consistency projection, $R$ is a deterministic renderer, and $G_{\phi}$ is an image-blind rewriter.

Every content item in $y_0$ is generated from a tuple in $\widetilde{E}$. The rewriter is instructed to preserve finding identity, polarity, and FDI number while altering ordering, aggregation, syntax, and lexical choice. Because learned realization is not guaranteed to be lossless, retaining $y_0$ makes rewrite-induced deviations directly auditable and allows evidence prediction to be evaluated separately from language realization.

%% file: sections/4_method.tex
\section{Method}

\subsection{Overview}
As illustrated in Figure~\ref{fig:pipeline}, the method isolates these measurable interfaces: anatomical localization, report-derived evidence recovery, deterministic consistency correction, and textual realization of the selected evidence. Stage~1 supplies anatomical masks and reusable structure tokens. Stage~2 maps anatomy-aligned inputs to a partially observed multilabel record. Stage~3 converts the record to prose while keeping the content interface discrete.

\subsection{Reference Evidence Construction}
Before model training, we construct reference evidence from each observed report with a two-pass, dental-knowledge-guided workflow. The first pass GPT-5.5 agent identifies coverage statements, tooth-indexed findings, tooth-IAC relations, global clinical facts, and normal or explicitly negative findings, while retaining the supporting report span for every item. A review-and-correction agent implemented with Claude Opus 4.8 then verifies each structured item against its source text, confirming 98.3\% on the first pass and directly correcting the remaining 1.7\% before Stage~2 label construction. Schema definitions and verification procedures are provided in Sect.~B of supplementary.

The resulting JSON is deliberately richer than the compact Stage~2 interface: it combines normalized entries with free-form facts and source text, whereas Stage~2 projects recurrent, learnable content into 11 tooth, 12 global, and 8 relation categories. The complete prediction schema and label frequencies are listed
in supplementary Sect.~B. The workflow structures the observed report; it does not claim to adjudicate the report or create exhaustive clinical truth.

\subsection{Stage 1: Anatomy Segmentation and Feature Exposure}
Stage~1 is an nnU-Net-v2-style 3D U-Mamba2 segmenter \citep{tan2025umamba2} trained on ToothFairy3 to predict 46 segmentation classes, including FDI-position tooth masks, mandible, maxilla, sinuses, canals, and dental-material masks. It supplies anatomical masks for crop construction and semantic channels, together with pooled encoder tokens for individual structures. The segmenter achieves a macro Dice of 0.899 over present structures and a case-averaged Dice of 0.880; training and feature-extraction details are provided in Sect.~C of supplementary.

\subsection{Stage 2: Reliability-Aware Evidence Prediction}
\paragraph{Branch-specific views.}
The schema is decomposed into three multilabel branches with separate anatomical inputs and classifiers. The tooth branch predicts 11 findings at each of 32 FDI positions from tooth-centered crops. The global branch predicts 12 volume-level findings from a down-sampled full volume. The relation branch predicts eight canal-related findings from left and right posterior-mandible views. This decomposition aligns each receptive field and output space with the corresponding clinical unit.

\paragraph{Anatomy-aligned fusion.}
Each branch contains a from-scratch 3D encoder that produces an appearance feature $r$. The global branch uses $r$ directly. The relation and tooth branches project the corresponding Stage~1 token into a context feature $c$ and use an anatomical embedding $s$ to identify the tooth or region. Their gated hybrid representation is
\begin{equation}
g=\sigma(W_g[r;c;s]+b_g), \qquad z=(1-g)\odot r+g\odot c.
\end{equation}
where \(W_g\) and \(b_g\) are the learnable projection matrix and bias of the element-wise fusion gate. The hybrid representation can exploit segmentation context while retaining raw intensity patterns that Stage~1 may discard. The main system uses a scratch global branch and a hybrid relation branch. The corresponding global-relation interface sweep is reported in supplementary Sect.~F.

\paragraph{Metal-sensitive intensity channel.}
Standard preprocessing compresses high-intensity CBCT values, obscuring differences among metallic restorations and prostheses. We therefore add a raw-intensity channel to each tooth crop, with its window estimated from high-density dental-material regions within the tooth-centered crop. It suppresses lower-density tissue while preserving normalized variation among high-density materials and artifacts, complementing the conventional anatomical channel.

\paragraph{Reliability-aware tooth-level non-mention supervision.}
Report parsing assigns each tooth-class pair an observed positive, an observed negative, or an unmentioned state. We apply non-mention supervision only to eligible tooth-level labels; unreported global and tooth-IAC labels remain unknown and are excluded from the corresponding losses. Motivated by the hypothesis that eligible tooth-level non-mentions are predominantly negative in aggregate, while remaining uncertain at the individual-case level, we compare three policies: \texttt{explicit} excludes it from the loss, \texttt{hard} assigns it a unit-weight negative label, and \texttt{pu-weak} assigns it a negative label with reduced, class-dependent weight. This reliability-weighting strategy is inspired by positive-unlabeled (PU) learning~\citep{kiryo2017pu}. The weighted tooth-branch objective is
\begin{equation}
\mathcal{L}=\sum_{i,c}\beta_c\,w(s_{ic},c)\,
\operatorname{BCE}(\widetilde{y}_{ic},p_{ic}),
\end{equation}
where \(\beta_c\) addresses class prevalence and \(w\) represents supervision reliability. Explicit positive and negative statements receive unit weight, whereas eligible tooth-level non-mentions receive \(0<w(\mathrm{unmentioned},c)<1\) under \texttt{pu-weak}. The schema-specific supervision rules and class-dependent weights are detailed in supplementary Sect.~B.

\paragraph{Tooth-level sampling and missingness conditioning.}
Within the tooth branch, per-class epoch sampling retains all supervised positives while limiting exposure to abundant negatives. A detached missing-tooth probability conditions the remaining attribute head, allowing it to distinguish present and likely-absent tooth contexts without propagating attribute loss into missingness prediction. The sampling and conditioning details are given in Sect.~C of supplementary.


\subsection{Stage 3: Evidence-Bottlenecked Report Realization}

\paragraph{Dental-logic consistency projection.} Before rendering, deterministic constraints derived from dental anatomical semantics map $\widehat{E}$ to $\widetilde{E}=\Pi_{\mathrm{dent}}(\widehat{E})$. Rather than treating selected labels independently, the projection removes combinations that violate predefined dental compatibility constraints, using predicted confidence only to resolve mutually exclusive alternatives. It operates solely on predicted tuples, introduces no positive finding, and leaves the corrected record inspectable at the evidence interface. The complete constraint definitions are provided in supplementary Sect.~C.

\paragraph{Deterministic evidence rendering.} A non-parametric renderer converts the consistency-projected evidence set \(\widetilde{E}\) into a canonical draft. Findings are grouped by compatible anatomical region and finding type, with compact templates used for repetitive patterns such as a fully edentulous arch. The resulting draft explicitly enumerates the content selected by Stage~2 while leaving sentence structure, ordering, aggregation, and clinical phrasing to the learned realizer.

\paragraph{Content-preserving teacher rewrites.} For each training case, we construct the draft from consistency-projected predicted evidence, thereby exposing the realizer to inputs resembling those encountered at deployment. GPT-4o \citep{openai2024gpt4o} then rewrites the draft into natural clinical language. The corresponding clinical report is provided only as a linguistic reference, and GPT-4o is instructed to preserve the factual content of the draft without adding, removing, or correcting findings. The resulting rewrites serve as realization targets rather than corrected evidence labels.

\paragraph{Image-blind behavioral distillation.} We distill the teacher's evidence-to-report realization behavior into a local Qwen2.5-7B rewriter adapted with low-rank adaptation (LoRA)~\citep{hu2022lora}. The student receives only the deterministic draft during both training and inference; it has no access to the CBCT volume, the reference report, or a parallel visual representation through which it could bypass the evidence bottleneck. It is therefore not explicitly trained to identify or correct Stage~2 prediction errors. We retain the deterministic draft for direct comparison, and inference requires no external model call. Decoding details are reported in supplementary Sect.~C. We empirically evaluate the trained student rewriter's content preservation and rewrite fidelity in \textit{Auditability and Rewrite Fidelity}, with additional analyses provided in supplementary Sect.~D.

%% file: sections/5_experiments.tex
\section{Experiments}

\begin{figure*}[t]
    \centering
    \includegraphics[width=0.8\textwidth]{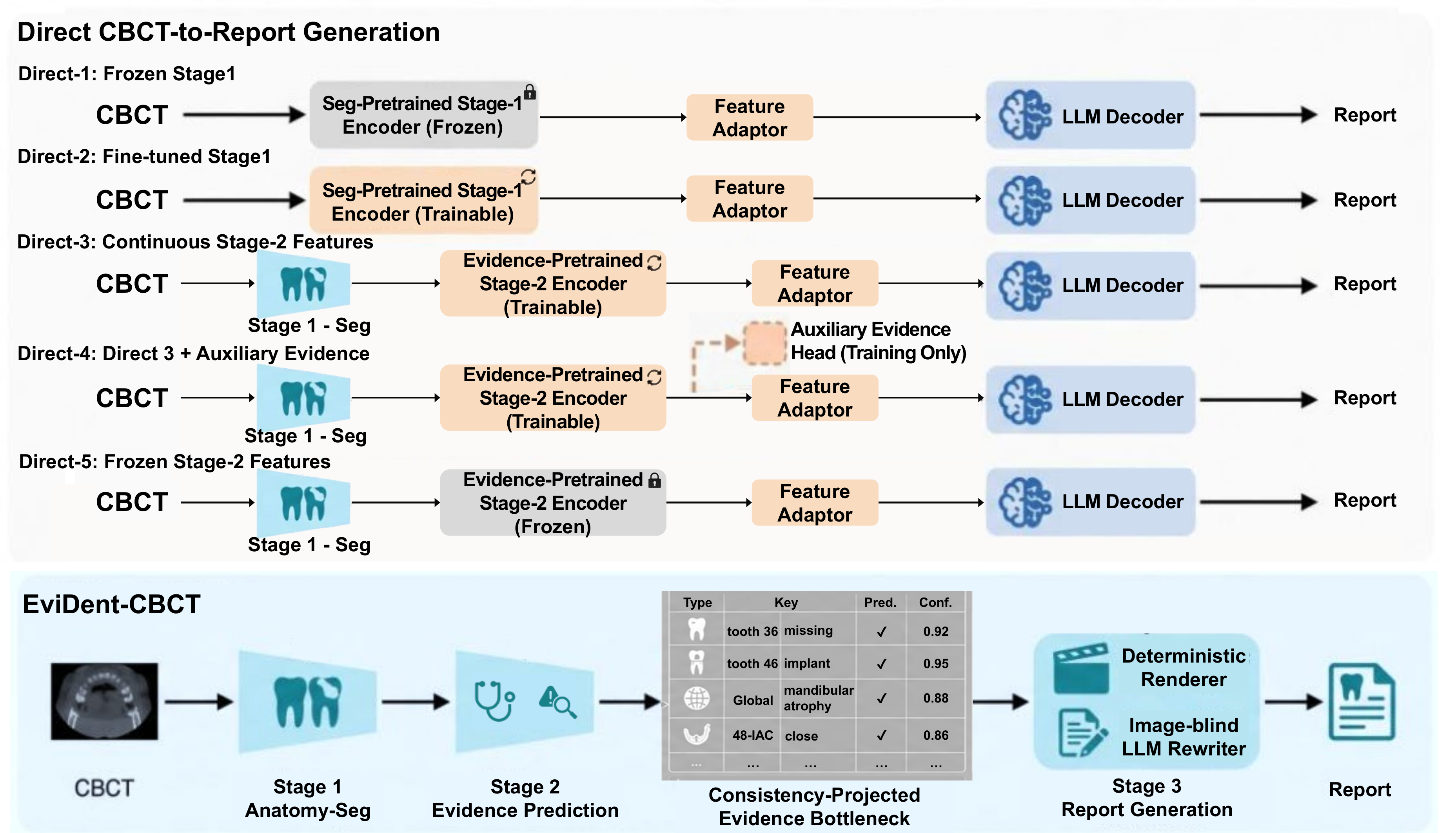}
    \caption{\textbf{Comparison of the inference-time interfaces used by the controlled direct generators and EviDent-CBCT.} Direct 1 and Direct 2 condition a language decoder on frozen or fine-tuned Stage~1 representations. Direct 3 fine-tunes Stage~2-checkpoint-initialized branches and supplies their continuous features to the decoder; Direct 4 adds training-only auxiliary evidence supervision, while Direct 5 freezes the Stage~2 branches. These five interfaces are instantiated with Qwen3-8B and Med42-8B decoders. EviDent-CBCT instead supplies a consistency-projected evidence record as the only case-specific input to an image-blind realizer. Its deterministic draft exposes the predicted findings and makes rewrite-induced deviations auditable.}
    \label{fig:conceptual-comparison}
\end{figure*}

\begin{table*}[t]
\centering
\scriptsize
\setlength{\tabcolsep}{1.7mm}
\begin{tabular}{@{}lllccccc@{}}
\toprule
Family & Visual/report interface & Report target & Logical P & Logical R & Logical F1 & BLEU-4 & METEOR \\
\midrule
Direct Qwen3-8B & Direct 1: frozen Stage~1 tokens & Raw & \(.216\pm.016\) & \(.245\pm.036\) & \(.229\pm.018\) & \(.121\pm.005\) & \(.352\pm.004\) \\
 & Direct 1: frozen Stage~1 tokens & Canonical & \(.295\pm.033\) & \(.294\pm.043\) & \(.294\pm.034\) & \(.095\pm.004\) & \(.349\pm.006\) \\
 & Direct 2: fine-tuned Stage~1 encoder & Raw & \(.257\pm.027\) & \(.251\pm.009\) & \(.253\pm.011\) & \(.135\pm.007\) & \(.338\pm.012\) \\
 & Direct 2: fine-tuned Stage~1 encoder & Canonical & \(.331\pm.056\) & \(.315\pm.023\) & \(.322\pm.036\) & \(.093\pm.007\) & \(.355\pm.005\) \\
 & Direct 3: continuous Stage~2 features & Raw & \(.238\pm.034\) & \(.236\pm.030\) & \(.236\pm.025\) & \(.124\pm.015\) & \(.325\pm.024\) \\
 & Direct 3: continuous Stage~2 features & Canonical & \(.351\pm.004\) & \(.314\pm.009\) & \(.331\pm.005\) & \(.099\pm.003\) & \(.356\pm.005\) \\
 & Direct 4: Direct 3 + auxiliary evidence & Raw & \(.240\pm.038\) & \(.249\pm.030\) & \(.244\pm.034\) & \(.119\pm.015\) & \(.322\pm.007\) \\
 & Direct 4: Direct 3 + auxiliary evidence & Canonical & \(.342\pm.008\) & \(.322\pm.027\) & \(.331\pm.014\) & \(.098\pm.003\) & \(.357\pm.015\) \\
 & Direct 5: frozen Stage~2 encoder & Raw & \(.239\pm.035\) & \(.233\pm.017\) & \(.234\pm.010\) & \(.121\pm.014\) & \(.326\pm.015\) \\
 & Direct 5: frozen Stage~2 encoder & Canonical & \(.346\pm.018\) & \(.311\pm.020\) & \(.327\pm.003\) & \(.097\pm.002\) & \(.358\pm.011\) \\
\midrule
Direct Med42-8B & Direct 1: frozen Stage~1 tokens & Raw & \(.227\pm.016\) & \(.264\pm.015\) & \(.243\pm.005\) & \(.121\pm.015\) & \(.360\pm.006\) \\
 & Direct 1: frozen Stage~1 tokens & Canonical & \(.356\pm.020\) & \(.334\pm.014\) & \(.344\pm.012\) & \(.098\pm.002\) & \(.353\pm.007\) \\
 & Direct 2: fine-tuned Stage~1 encoder & Raw & \(.266\pm.031\) & \(.293\pm.016\) & \(.278\pm.024\) & \(.129\pm.009\) & \(.352\pm.009\) \\
 & Direct 2: fine-tuned Stage~1 encoder & Canonical & \(\mathbf{.388\pm.019}\) & \(.354\pm.017\) & \(.371\pm.018\) & \(.100\pm.002\) & \(.360\pm.003\) \\
 & Direct 3: continuous Stage~2 features & Raw & \(.291\pm.003\) & \(.272\pm.036\) & \(.280\pm.018\) & \(.134\pm.012\) & \(.341\pm.011\) \\
 & Direct 3: continuous Stage~2 features & Canonical & \(.366\pm.005\) & \(.359\pm.011\) & \(.362\pm.006\) & \(.099\pm.001\) & \(.357\pm.003\) \\
 & Direct 4: Direct 3 + auxiliary evidence & Raw & \(.273\pm.020\) & \(.272\pm.019\) & \(.271\pm.002\) & \(.127\pm.017\) & \(.324\pm.017\) \\
 & Direct 4: Direct 3 + auxiliary evidence & Canonical & \(.374\pm.008\) & \(.356\pm.010\) & \(.365\pm.007\) & \(.101\pm.002\) & \(.361\pm.004\) \\
 & Direct 5: frozen Stage~2 encoder & Raw & \(.252\pm.004\) & \(.274\pm.020\) & \(.262\pm.010\) & \(.120\pm.013\) & \(.313\pm.011\) \\
 & Direct 5: frozen Stage~2 encoder & Canonical & \(.380\pm.024\) & \(.342\pm.006\) & \(.360\pm.013\) & \(.099\pm.001\) & \(.354\pm.003\) \\
\midrule
EviDent-CBCT & Consistency-projected evidence record & Deterministic template & \(.372\pm.002\) & \(\mathbf{.428\pm.008}\) & \(.398\pm.004\) & \(\mathbf{.158\pm.003}\) & \(\mathbf{.425\pm.004}\) \\
 & Consistency-projected evidence record & Local distilled rewrite & \(.384\pm.002\) & \(.423\pm.004\) & \(\mathbf{.402\pm.003}\) & \(.083\pm.001\) & \(.386\pm.001\) \\
\bottomrule
\end{tabular}
\caption{Validation set results, reported as mean\(\pm\)std over three runs. Logical metrics use RadFact-Lite-Dental.}
\label{tab:main}
\end{table*}

\subsection{Dataset}

The paired subset of ToothFairy4 contains 624 CBCT volumes and 1,001 English reports~\citep{lumetti2026ontology}. Reports associated with the same volume are consolidated at the label level; when they disagree on a binary finding, we exclude that label from supervision rather than arbitrate between readers. As each scan contributes many tooth-level instances, this conservative treatment still provides ample, unambiguous supervision for Stage 2, while modeling inter-report disagreement remains future work. We split the 624 paired volumes into 561 training and 63 validation cases. The same volume-level partition was applied to all three stages and all direct baselines. Any ToothFairy3 volume corresponding to a validation case was excluded from Stage~1 training. ToothFairy4 additionally includes a hidden test set of 50 CBCT scans from an independent center, evaluated by the challenge organizers; official results are reported below.

\subsection{Evaluation Metrics}

\paragraph{Merged Stage~2 logical set-F1.}
For each case, positive tuples parsed from the reference report are pooled across the tooth, global, and relation branches within the evaluation mask. Predicted and reference-derived tuples match only when their structured keys are identical. We compute per-case precision and recall, average each quantity over cases, and take the harmonic mean of those two averages. This RadFact-style aggregation is an LLM-free measure of structured agreement with the reports. Merged Stage~2 logical set-F1 is computed on the selected tuples before consistency projection, isolating the learned evidence predictor from deterministic correction. Its formal definition and aggregation conventions are given in supplementary Sect.~G.

\paragraph{Internal report metrics.}
RadFact decomposes each narrative report into clinical propositions and evaluates entailment bidirectionally~\citep{bannur2024maira2}. Candidate propositions supported by the reference determine logical precision, while reference propositions supported by the candidate determine logical recall; logical F1 is their harmonic mean. We build on RadFact-Lite~\citep{radfactlite010} with a dentist-guided dental-atomic parser, FDI normalization, and dental entailment instructions. Exact and rule-guaranteed matches are resolved deterministically, while comparisons are judged by GPT-5.6-Luna. The original bidirectional aggregation is unchanged. We report these metrics together with BLEU-4~\citep{papineni2002bleu} and METEOR~\citep{banerjee2005meteor}. Main-paper metrics use the same fixed reference report per volume; supplementary Sect.~E provides a mean-over-references robustness analysis. Full details and an author audit of RadFact-Lite-Dental are provided in supplementary Sect.~G. The official ODIN evaluation uses a different RadFact protocol. We therefore report its scores separately and do not directly compare their absolute values with RadFact-Lite-Dental.

\paragraph{Paired uncertainty analysis.}
For the primary comparison, we average case-level logical metrics across runs, jointly bootstrap matched cases, recompute corpus-level logical F1, and report the percentile 95\% confidence interval of paired differences. Details are provided in supplementary Sect.~E.

\vspace{-1mm}
\subsection{EviDent-CBCT vs. Controlled Direct Generation}


Figure~\ref{fig:conceptual-comparison} summarizes EviDent-CBCT and five controlled direct-generation interfaces. Direct~1 projects frozen Stage~1 structure tokens, while Direct~2 recomputes these tokens from a fixed-size volume and fine-tunes the encoder levels that produce them. Direct~3 initializes the global, relation, and tooth branches from the corresponding Stage~2 checkpoints and supplies their trainable continuous features to the language decoder; Direct~4 additionally applies Stage~2 matched auxiliary supervision; and Direct~5 freezes the Stage~2 branches while retaining a trainable feature adapter and decoder, testing whether direct SFT degrades the evidence-pretrained representation. Each interface is instantiated with either Qwen3-8B~\citep{yang2025qwen3} or Med42-8B~\citep{christophe2024med42} and trained with both raw and canonical report targets. The \emph{raw} target is the observed clinical report, whereas the \emph{canonical} target is obtained by rendering reference-derived evidence and passing the draft through the frozen rewriter, thereby reducing linguistic and content-selection variability.
Further implementation details and qualitative comparisons are provided in supplementary Sect.~E and Sect.~I, respectively.

Canonical targets improve every decoder-interface pair. Med42 strengthens the direct controls, with canonical Direct 2 reaching \(.371\pm.018\) F1, compared with at most \(.331\) for Qwen3. Among Stage~2-initialized Med42 controls, Direct 3 and Direct 4 reach \(.362\pm.006\) and \(.365\pm.007\), respectively. Freezing the Stage~2 branches in Direct 5 yields a decline with both Qwen3 and Med42, providing no improvement over the corresponding trainable Direct 3 pathways.

EviDent-CBCT reaches \(.398\pm.004\) logical-F1 with deterministic templates and \(.402\pm.003\) after local rewriting. The template also obtains the highest BLEU-4 and METEOR, showing that the comparison does not depend on the learned realizer. Against the strongest Med42 direct configuration, the rewritten output has an unrounded paired F1 gain of .0317, with a 95\% bootstrap confidence interval of [.0047, .0586]. Moreover, Direct 3 and Direct 4 use the same evidence-pretrained Stage~2 initialization while retaining a continuous, bypassable interface. These controls support making the discrete evidence record the required report interface rather than only a pretraining signal or auxiliary target. Per-run results and specifications are reported in supplementary Sect.~E. This advantage persists under mean-over-references evaluation (.374 vs.\ .333; supplementary Sect.~E).


\subsection{Stage-wise Analysis of Evidence-to-Report Pathway}

Table~\ref{tab:ceiling} traces the successive representations from the observed report to the generated output. Reference self-comparison reaches 1.000 logical-F1. The extracted source spans cover 96.5\% of alphanumeric report content on average, and their verbatim reassembly reaches .957 logical-F1. Together with the second-agent verification pass, this indicates that the extraction workflow retains most evaluator-recognized report content. The residual consists mainly of sparse orientation notes, classification terms, and broad normal statements; detailed coverage analysis is provided in supplementary Sect.~B.

The largest loss occurs when the richer evidence JSON is projected into the compact prediction schema and deterministically rendered, reducing logical-F1 from .957 to .589. The schema preserves recurrent finding identities but omits details without dedicated labels, including the distribution and degree of atrophy, treatment quality, bridge topology, and lesion morphology or size. Qwen realization raises the reference-derived score to .597 by restoring narrative organization, aggregation, and clinically familiar phrasing. Because the rewriter receives only the deterministic draft, .597 is an operational ceiling for the current compact schema-to-report pathway, not for the richer evidence JSON.


We next quantify the effect of the Stage~3 dental-logic consistency projection. Its constraints were fixed before the final report-level evaluation and modified \(178\) tuple instances across \(52\) unique validation cases over three runs: \(162\) missing-tooth/IAC conflicts and \(16\) contradictory canal-course predictions. The projection raises template F1 from \(.389\pm.006\) to \(.398\pm.004\) and rewritten F1 from \(.398\pm.001\) to \(.402\pm.003\), showing that correction occurs at the evidence interface rather than in the language realizer. The remaining gap to the reference-derived pathway primarily reflects upstream evidence prediction.

\begin{table}[t]
\centering
\small
\setlength{\tabcolsep}{0.3mm}
\begin{tabular}{@{}lc@{}}
\toprule
Candidate report & Logical F1 \\
\midrule
Reference report vs. itself & 1.000 \\
Verbatim source clues from evidence extraction & .957 \\
Reference-derived evidence $\rightarrow$ template & .589 \\
Reference-derived evidence $\rightarrow$ template $\rightarrow$ Qwen & .597 \\
Predicted evidence $\rightarrow$ template & \(.389\pm.006\) \\
Predicted evidence $\rightarrow$ template $\rightarrow$ Qwen & \(.398\pm.001\) \\
Consistency-projected evidence $\rightarrow$ template & \(.398\pm.004\) \\
Consistency-projected evidence $\rightarrow$ template $\rightarrow$ Qwen & \(.402\pm.003\) \\
\bottomrule
\end{tabular}
\caption{Evidence-to-report pathway decomposition and
reference-derived operational ceiling. Reference-derived rows are evaluated once; predicted-evidence rows are reported as mean$\pm$std over three runs.}
\label{tab:ceiling}
\end{table}

\begin{table}[t]
\centering
\scriptsize
\setlength{\tabcolsep}{0.5mm}
\begin{tabular}{@{}llccc@{}}
\toprule
Axis & Variant & Logical P & Logical R & Logical F1 \\
\midrule
Base & gated, metal, \texttt{pu-weak} & \(.434\pm.028\) & \(.538\pm.022\) & \(\mathbf{.480\pm.007}\) \\
\midrule
Supervision & \texttt{hard} & \(.408\pm.017\) & \(\mathbf{.548\pm.030}\) & \(.467\pm.017\) \\
 & \texttt{explicit} & \(.405\pm.057\) & \(.328\pm.035\) & \(.359\pm.018\) \\
\midrule
Fusion & concat & \(\mathbf{.435\pm.036}\) & \(.525\pm.033\) & \(.474\pm.015\) \\
 & visual-only & \(.422\pm.018\) & \(.534\pm.033\) & \(.471\pm.004\) \\
 & scratch-only & \(.379\pm.021\) & \(.512\pm.039\) & \(.435\pm.001\) \\
\midrule
Metal & no metal channel & \(.415\pm.019\) & \(.527\pm.039\) & \(.463\pm.006\) \\
\bottomrule
\end{tabular}%
\caption{Tooth-branch Stage~2 logical set metrics, reported as mean\(\pm\)std over three model runs. All non-ablated branches and inputs are held fixed.}
\label{tab:tooth-ablation}
\end{table}

\vspace{-1mm}
\subsection{Auditability and Rewrite Fidelity}
The deterministic draft makes learned realization auditable at the proposition level. An author reviewed 30 evaluation cases for additions, omissions, polarity changes, FDI or anatomical-side reassignment, finding substitutions, and redundancy. All template propositions were preserved in 28 cases (93.3\%), with no polarity reversal, FDI reassignment, or anatomical-side reassignment. The two substantive changes both aligned the rewrite more closely with the reference: one removed a crown prediction where the reference described a restoration, and the other changed a restoration to a crown. Rewriting is therefore not guaranteed to be lossless, even though individual changes can regularize imperfect predicted evidence. Its modest aggregate improvement is reported in Table~\ref{tab:ceiling}, and retaining the pre-rewrite draft keeps case-level changes inspectable. The full audit protocol and case-level deviations are reported in supplementary Sect.~D.

\subsection{Stage~2 Evidence Learning and Tooth Ablations}
The final \texttt{g\_scratch\_r\_hybrid} pipeline uses a scratch global branch, a hybrid relation branch, and the tooth-base configuration. Across three runs, it reaches \(.666\pm.006\) merged Stage~2 logical set-F1 and \(.402\pm.003\) final report logical-F1. Tooth evidence is the most fine-grained component of the merged record and the focus of Table~\ref{tab:tooth-ablation}. Supplementary Sect.~H further reports class-wise recognition results for the tooth branch under \texttt{pu-weak}, \texttt{hard}, and \texttt{explicit} supervision, together with class-wise results for the fixed global and relation branches.

Within the tooth branch, treating every eligible non-mention as a full-weight negative (\texttt{hard}) raises logical set-F1 from \(.359\pm.018\) with \texttt{explicit} supervision to \(.467\pm.017\), primarily by increasing recall from \(.328\) to \(.548\). This suggests that tooth-level report silence provides useful negative signal in aggregate. Assigning class-dependent reduced weights to these non-mentions (\texttt{pu-weak}) gives the best balance, with \(.434\pm.028\) precision, \(.538\pm.022\) recall, and \(.480\pm.007\) F1. Within the current split and weighting scheme, these results support reduced-weight supervision for eligible tooth-level non-mentions, but do not establish that the selected weights are clinically calibrated or that the same policy applies to global and relation labels.

The fusion variants are close: gated hybrid, concat, and visual-only reach tooth-branch F1 values of \(.480\), \(.474\), and \(.471\), whereas scratch-only falls to \(.435\). Removing the metal-sensitive channel lowers tooth-branch F1 from \(.480\) to \(.463\). These ablations are measured directly at the Stage~2 tooth interface, with the global and relation branches held fixed.

\subsection{Official ODIN 2026 Hidden-Test Evaluation}

On the ODIN 2026 hidden test set\footnote{\url{https://odin2026.grand-challenge.org/challenge-winners/}}, EviDent-CBCT achieved an official aggregate score of 0.3033 (RadFact F1 0.3245, BLEU-4 0.1206, METEOR 0.3171), ranking second in automated evaluation and third in the clinical Arena. The official RadFact protocol differs from our RadFact-Lite-Dental evaluation, so the absolute scores are not directly comparable.

\begin{table*}[t]
\centering
\caption{Official ODIN 2026 ToothFairy4 hidden-test results for the top three teams.
The official RadFact protocol differs from RadFact-Lite-Dental used in our internal validation.}
\label{tab:odin_hidden}
\begin{tabular}{lcccccccc}
\toprule
Team & RadFact F1 & BLEU-4 & METEOR & Aggregate Score & Arena Score \\
\midrule
shivam.singh & 0.3924 & 0.1476 & 0.3269 & 0.3614 & 1307 \\
\textbf{EviDent-CBCT(ours)} & 0.3245 & 0.1206 & 0.3171 & 0.3033 & 985 \\
ViktorB & 0.2321 & 0.0865 & 0.2828 & 0.2226 & 1172 \\
\bottomrule
\end{tabular}
\end{table*}

%% file: sections/6_conclusion.tex
\section{Conclusion and Outlook}
We presented EviDent-CBCT, an evidence-bottlenecked framework for learning dento-maxillofacial CBCT reports from non-exhaustive routine-report supervision. Its discrete record separates evidence prediction from language realization, enables deterministic dental-consistency correction, and makes selected report content auditable. For tooth-level evidence, eligible non-mentions receive reduced-weight negative supervision, while unreported global and tooth-IAC labels remain unknown. Across three validation runs, the system achieves \(0.402\pm0.003\) RadFact-Lite-Dental logical-F1, compared with \(0.371\pm0.018\) for the strongest controlled direct generator. It further ranked second in the ODIN 2026 automated evaluation and third in the blinded clinical comparison. The controlled comparisons, supervision ablations, and coverage analysis support the discrete evidence record as a structured content interface for report-grounded CBCT generation.

These findings remain limited by non-exhaustive report supervision, the lack of independently adjudicated image-level truth, and a compact schema that underrepresents rare qualifiers and findings. The hidden-test evaluation provides independent-center testing and blinded clinical assessment, although the test references remain unavailable for detailed error analysis. Future work will model cross-report disagreement, preserve rare qualifiers, and evaluate additional external cohorts with blinded dental readers.

%% file: sections/7_supplementary.tex
\clearpage
\appendix

\twocolumn[
\begin{center}
    {\LARGE\bfseries Supplementary Material of EviDent-CBCT\par}
\end{center}
\vspace{0.8em}
]

\section{Scope of the Report-Derived Supervision}
EviDent-CBCT starts from the clinical reports available in ToothFairy4. We do not interpret these reports as exhaustive image annotations or construct an independently adjudicated medical gold standard. A non-mention may reflect true absence, report-specific omission, habitual non-reporting, oversight, or disagreement across readers. These mechanisms are not adjudicated and cannot be distinguished from the observed reports alone. Explicit report inaccuracies are also possible, but their prevalence is not measured in this study.

The three policies in the main paper apply only to eligible tooth-level non-mentions. \texttt{explicit} excludes these entries from the loss, \texttt{hard} assigns them unit negative weight, and \texttt{pu-weak} assigns class-dependent reduced negative weights. Unreported global and tooth-IAC categories remain unknown and are excluded from supervision because their positive labels are sparse. None of these policies adjudicates explicit report inaccuracies, and our experiments do not estimate their prevalence. Accordingly, the reported Stage~2 metric measures structured agreement with the reports rather than clinical accuracy against an external standard.

\section{Reference Evidence Construction and Coverage Audit}
\subsection{Construction and independent verification}
The full reference evidence JSON is produced before Stage~2 preprocessing. GPT-5.5 (medium effort) receives the original report and a dental-specific extraction prompt covering FDI notation, arch and quadrant expressions, missing and treated teeth, prostheses and implants, bone findings, scan coverage, and tooth-IAC relations. It returns structured sections for coverage, tooth labels, relation labels, additional clinical facts, and normal or explicitly negative statements. Every item includes the report text that supports it, so the normalized entry remains traceable to the observed wording.

A second agent implemented through Claude Opus4.8 independently reviews each JSON against the report. The reviewer checks support, polarity, tooth identity, anatomical assignment, completeness, and the correspondence of each attached span; it either accepts the entry or requests calibration. The reported 98.3\% first-pass confirmation rate is the proportion of entries that this Opus4.8 reviewer accepted without requesting calibration. The remaining 1.7\% are flagged and corrected before constructing Stage~2 labels. This is agent-review agreement with the report, not dentist validation or correctness against an independently read CBCT volume.

\subsection{Full JSON versus the prediction schema}
Two representations should not be conflated. The full evidence JSON preserves categorical fields, free-form facts, normal/negative statements, and verbatim source evidence. The Stage~2 interface is a task-oriented projection containing 11 tooth-level, 12 global, and eight relation categories. Rare details can therefore remain auditable in the full JSON without becoming dedicated output classes for image-based prediction. The ceiling row ``verbatim source clues'' evaluates the former; the template and Qwen ceiling rows evaluate the latter after projection and realization.

\subsection{Complete Stage~2 evidence schema}
Tables~\ref{tab:schema-tooth}-\ref{tab:schema-relation} enumerate the complete prediction schema. Counts are positive reference-derived labels pooled across the fixed training and validation partitions. Tooth counts denote FDI-position-category instances; global and relation counts denote case-category instances. They are label frequencies in the observed reports, not estimates of clinical prevalence. For sparse global and relation labels, an unreported category remains unknown. For tooth labels, non-mentions eligible for positive-unlabeled supervision are assigned the reliability weights in Table~\ref{tab:schema-tooth}; explicit positive and negative statements receive weight 1, while unavailable positions and contradictory statements receive weight 0.

\begin{table*}[t]
\centering
\small
\begin{tabular}{@{}p{0.25\textwidth}p{0.34\textwidth}rrr@{}}
\toprule
Schema field & Meaning & Total $n_+$ & Observed $w_-$ & Report-level $w_-$ \\
\midrule
\path{missing} & Tooth absent at the indexed FDI position & 924 & .95 & .95 \\
\path{implant} & Endosseous implant at the indexed position & 256 & .80 & .50 \\
\path{crown} & Prosthetic crown at the indexed position & 695 & .60 & .40 \\
\path{restoration} & Coronal restoration or filling, distinct from a crown & 1,792 & .50 & .30 \\
\path{endodontic_treatment} & Root-canal or other endodontic treatment & 766 & .70 & .40 \\
\path{prosthesis_related} & Bridge, pontic, or other prosthetic involvement & 1,305 & .50 & .30 \\
\path{impacted_or_disodontiasis} & Impaction or disodontiasis & 365 & .80 & .50 \\
\path{caries} & Dental caries & 233 & .20 & .10 \\
\path{root_remnant} & Retained root or root remnant & 61 & .70 & .40 \\
\path{root_fracture} & Root fracture & 11 & .20 & .10 \\
\path{lesion_related} & Tooth-associated lesion finding & 388 & .30 & .15 \\
\bottomrule
\end{tabular}
\caption{Complete tooth-level prediction schema, positive sample counts, and class-dependent \texttt{pu-weak} weights. \emph{Observed} denotes a non-mention for a tooth observed elsewhere in the report; \emph{report-level} denotes an otherwise eligible non-mention inferred only from report silence.}
\label{tab:schema-tooth}
\end{table*}

\begin{table*}[t]
\centering
\small
\begin{tabular}{@{}p{0.32\textwidth}p{0.56\textwidth}r@{}}
\toprule
Schema field & Meaning & Total $n_+$ \\
\midrule
\path{maxilla_partially_included} & Maxilla only partly included in the scan field & 236 \\
\path{maxilla_not_included} & Maxilla excluded from the scan field & 234 \\
\path{maxillary_sinuses_not_included} & Maxillary sinuses excluded from the scan field & 173 \\
\path{mandibular_condyles_not_included} & Mandibular condyles excluded from the scan field & 588 \\
\path{periodontitis_present} & Reported periodontal disease & 207 \\
\path{atrophy_maxilla} & Maxillary alveolar atrophy & 53 \\
\path{atrophy_mandible} & Mandibular alveolar atrophy & 312 \\
\path{atrophy_severe} & Severe or marked atrophy in either arch & 113 \\
\path{osteolytic_lesion_present} & Osteolytic or radiolucent osseous lesion & 200 \\
\path{osteosclerotic_lesion_present} & Osteosclerotic or osteocondensing lesion & 38 \\
\path{sinus_pathology_present} & Pathological finding in a maxillary sinus & 58 \\
\path{poor_quality_or_artifact} & Poor image quality or acquisition/reconstruction artifact & 22 \\
\bottomrule
\end{tabular}
\caption{Complete global prediction schema and positive sample counts.}
\label{tab:schema-global}
\end{table*}

\begin{table*}[t]
\centering
\small
\begin{tabular}{@{}p{0.36\textwidth}p{0.52\textwidth}r@{}}
\toprule
Schema field & Meaning & Total $n_+$ \\
\midrule
\path{38_close_to_left_IAC} & Tooth 38 close to the left IAC & 168 \\
\path{48_close_to_right_IAC} & Tooth 48 close to the right IAC & 168 \\
\path{any_tooth_close_to_IAC} & Any reported tooth-IAC proximity & 219 \\
\path{lesion_close_to_IAC} & Lesion close to an IAC & 14 \\
\path{left_IAC_predominantly_lingual} & Predominantly lingual course of the left IAC & 236 \\
\path{left_IAC_predominantly_buccal} & Predominantly buccal course of the left IAC & 60 \\
\path{right_IAC_predominantly_lingual} & Predominantly lingual course of the right IAC & 243 \\
\path{right_IAC_predominantly_buccal} & Predominantly buccal course of the right IAC & 61 \\
\bottomrule
\end{tabular}
\caption{Complete relation prediction schema and positive sample counts. IAC denotes the inferior alveolar canal.}
\label{tab:schema-relation}
\end{table*}

\subsection{Coverage protocol}
For each evaluation case, we recursively collect the attached \texttt{text} fields from the coverage, tooth-label, relation-label, free-form fact, and normal/negative sections. After whitespace normalization, duplicate clues and shorter clues wholly contained in longer ones are removed, and the retained spans are restored to their order in the original report. This yields 555 unique clues. Every retained clue is an exact substring after whitespace normalization.

We report two complementary coverage measures. \emph{Literal coverage} is the fraction of alphanumeric report characters contained in the union of retained source spans; its case-averaged value is 96.5\% and its median is 100\%. \emph{Semantic coverage} concatenates the retained clues and compares them with the report using RadFact-Lite-Dental. The resulting logical precision, recall, and F1 are 0.986, 0.929, and 0.957, respectively. Literal coverage measures retained wording, whereas logical recall measures whether the selected spans support the propositions recovered from the report.

\subsection{Residual content}
We inspect report text outside the union of all retained source spans after excluding section headers, punctuation, and connective fragments. Table~\ref{tab:residual-content} summarizes the substantive residuals. They are concentrated in low-frequency metadata, classifications, and broad normal statements rather than recurring core dental findings.

\begin{table}[h]
\centering
\small
\begin{tabular}{@{}p{0.8\columnwidth}c@{}}
\toprule
Residual category & Cases \\
\midrule
Image orientation, laterality discrepancy, or acquisition note & 3 \\
Low-frequency dental classification (Kennedy class) & 1 \\
Explicit complete-dentition or complete-arch statement & 2 \\
Recurring omission of common dental findings & 0 \\
\bottomrule
\end{tabular}
\caption{Substantive report content not fully represented by retained evidence spans. Counts denote affected evaluation cases; categories may overlap within a case.}
\label{tab:residual-content}
\end{table}

The audit supports a precise interpretation: major report content is comprehensively captured, and the supporting text preserves most fine detail verbatim, but the normalized structured fields intentionally emphasize macroscopic, recurrent concepts. The residual concepts are too sparse in this corpus to justify separate supervised classes. The decrease from 0.957 for source clues to 0.597 for reference-derived evidence after template rendering and Qwen rewriting should therefore be attributed primarily to post-extraction projection and realization, rather than to wholesale failure of the initial evidence extraction.

\subsection{Coverage-to-ceiling interpretation}
The coverage audit and ceiling study examine successive representations of the same observed reports. Source spans retained in the full evidence JSON preserve fine-grained statements verbatim, reaching 96.5\% mean literal coverage and 0.957 logical-F1 when reassembled. Together with the two-pass verification results, this indicates that the extraction workflow captures the major content of the observed reports.

The largest decrease occurs when this rich representation is projected into the compact 31-category prediction schema and deterministically rendered, reducing logical-F1 from 0.957 to 0.589. This projection preserves recurrent findings but intentionally collapses details that are not modeled as dedicated labels, including the precise distribution and degree of atrophy, treatment quality, bridge topology, and lesion morphology or size. Rare free-form findings and normal statements outside the schema are likewise unavailable to the renderer. The decrease therefore reflects the combined granularity of the prediction schema and its deterministic realization, rather than a failure to identify those statements during initial extraction.

Qwen realization raises logical-F1 from 0.589 to 0.597 for reference-derived evidence by restoring narrative organization, aggregation, and clinically familiar phrasing. Because the rewriter receives only the deterministic draft, it cannot systematically recover information removed during schema projection. The 0.597 reference-derived result is therefore an operational ceiling for the compact schema-to-report pathway, not for the richer evidence JSON. Replacing reference-derived evidence with predicted evidence lowers the rewritten score to \(0.398\pm0.001\), isolating the remaining effect of Stage~2 evidence prediction. For predicted evidence, rewriting raises mean F1 from \(0.389\pm0.006\) to \(0.398\pm0.001\), primarily through improved precision.

\section{Implementation Details}
\paragraph{Stage~1.}
The 46-class nnU-Net-v2-style U-Mamba2 segmenter uses \(128\times256\times256\) patches and a 1,500-epoch schedule. In addition to segmentation masks, it exposes 1,408-dimensional pooled encoder tokens for individual teeth and anatomical structures.

\paragraph{Stage~2.}
The fixed tooth configuration performs label-wise sampling at the tooth-instance level. For every class and epoch, all supervised positives are retained and eligible negatives are sampled without replacement. The target negative-to-positive ratio is \(0.5\leq n_-/n_+<2\) when enough negatives are available; otherwise all available negatives are retained. Sampling determines the active label-instance pairs, while their explicit or \texttt{pu-weak} reliability weights remain in the loss. The \texttt{PU-weak} weights are predefined reliability weights under the guidance of the clinical routine, not learned or tuned on validation data. Unmentioned negatives receive class-dependent weights reflecting how reliably non-mention implies absence. For example, if a mask-present tooth is not reported as missing, it is highly likely to be present, so missing receives a high weak-negative weight. In contrast, non-mention of caries is less reliable because subtle caries may be overlooked or selectively omitted, so caries receives a lower weight.

For positions with a Stage~1 tooth mask, the mask center and principal axes define the oriented crop. Mask-absent positions receive only a fixed crop centered by interpolation from teeth in the same quadrant and no tooth-branch supervision; their missing status is inherited directly from the Stage~1 mask during branch merging. For a classifiable position with tooth feature \(u\), the missing probability \(m=\sigma(h_m(u))\) conditions the remaining attributes through \([u;\,m_{\mathrm{detach}}\phi(u)]\), allowing missingness to coexist with residual or prosthetic findings without propagating attribute loss into the missing head. Each branch is trained independently. The main configuration uses a scratch global branch, a hybrid relation branch, and the gated hybrid tooth branch with the metal-sensitive channel and \texttt{pu-weak} supervision.

Table~\ref{tab:stage2-input-encoders} specifies the actual tensors and encoders used by the main configuration. Dimensions and physical extents follow \((z,y,x)\). The global view covers the complete available scan and is resized to a fixed grid, so its effective spacing varies with the acquired field of view. For each tooth, PCA is applied to the 3D coordinates of its segmentation mask to define a tooth-centered local coordinate system; PCA is used only to estimate crop orientation, not for dimensionality reduction. An oriented bounding box enclosing the complete tooth is then expanded by 6 mm in all directions and resized to the fixed network input size. Because tooth dimensions vary across cases, this operation produces tooth-dependent effective voxel spacing. Relation crops instead use a fixed \(80\times96\times64\)-mm field of view; resizing it to \(80\times128\times128\) gives an effective resolution of \(1.0\times0.75\times0.5\) mm.

\begin{table*}[t]
\centering
\scriptsize
\begin{tabular}{@{}p{0.09\textwidth}p{0.21\textwidth}p{0.17\textwidth}p{0.45\textwidth}@{}}
\toprule
Branch & View and tensor size & Resolution/channels & Encoder and fusion \\
\midrule
Global & Whole available scan; \(96\times192\times192\) & Case-dependent effective spacing; 1 normalized CT + 5 semantic channels & Seven-stage 3D residual encoder with widths \([32,64,128,256,320,320,320]\), block counts \([1,3,4,6,6,6,6]\), and strides \([1,2,2,2,2,2,1]\); terminal 320-channel MambaLayer2, global average pooling, and a 12-label MLP. This branch is trained scratch-only. \\
Tooth & One PCA-aligned crop per FDI position; \(64\times64\times96\), with the tooth long axis mapped to \(x\) & Tooth-dependent spacing after adding 6 mm on each crop axis; 1 anatomical CT + 1 metal-sensitive CT + 8 semantic channels & SmallCropEncoder: three \(3 \times 3 \times 3\) convolutions with 12, 24, and 48 channels, instance normalization and LeakyReLU, pooling after the first two blocks, global pooling, and a 192-D projection. The crop feature is gated with a 192-D projection of the 1,408-D Stage~1 tooth token and a 16-D ten-region embedding, yielding a 256-D feature for label-specific heads; the missing head conditions the remaining attributes. \\
Relation & Paired left/right posterior-mandibular crops; each \(80\times128\times128\) from an \(80\times96\times64\)-mm FOV & \(1.0\times0.75\times0.5\) mm; 1 normalized CT + 5 semantic channels & The same 12-24-48 SmallCropEncoder produces a 192-D feature per side. Left/right features and 16-D side embeddings are combined with projected Stage~1 relation tokens by gated fusion into 384 dimensions, followed by an eight-label MLP. \\
\bottomrule
\end{tabular}
\caption{Stage~2 input geometry and encoder architecture in the primary configuration. CT and semantic-mask channels are concatenated before the branch encoder.}
\label{tab:stage2-input-encoders}
\end{table*}

\paragraph{Stage~3.}
Before rendering, the selected record is mapped through the deterministic dental-logic projection \(\widetilde{E}=\Pi_{\mathrm{dent}}(\widehat{E})\). A tooth-specific IAC-proximity tuple for position 38 or 48 is omitted when that position is marked missing, except when impaction or root-remnant evidence at the same position indicates residual dental structure. When predominantly lingual and predominantly buccal canal-course tuples are both selected on one side, the tuple with the higher predicted probability is retained. The projection uses only Stage~2 tuples and confidence values, and it never introduces a positive finding absent from \(\widehat{E}\). GPT-4o rewrites 561 training drafts constructed from consistency-projected predicted evidence. Qwen2.5-7B is adapted with LoRA rank 16 for three epochs using completion-only loss, and inference uses greedy decoding. The rewriter receives only the deterministic draft.

\section{Rewrite Content-Preservation Audit}
An author manually reviewed 30 evaluation cases selected by simple random sampling from the primary EviDent-CBCT run. For each case, the deterministic template and Qwen rewrite were compared proposition by proposition for additions, omissions, polarity changes, FDI reassignment, anatomical-side reassignment, finding-type substitutions, and redundant statements. The rewrite preserved all template propositions in 28 of 30 cases (93.3\%). Table~\ref{tab:rewrite-audit} summarizes the observed deviations.

\begin{table}[h]
\centering
\small
\begin{tabular}{@{}lc@{}}
\toprule
Audit outcome & Cases \\
\midrule
All template propositions preserved & 28 \\
Any substantive content change & 2 \\
Omission & 1 \\
Finding-type substitution & 1 \\
Polarity reversal & 0 \\
FDI reassignment & 0 \\
Anatomical-side reassignment & 0 \\
Redundant repetition & 1 \\
\bottomrule
\end{tabular}
\caption{Author audit of content changes introduced by Qwen rewriting. Categories can overlap within a case.}
\label{tab:rewrite-audit}
\end{table}

Both substantive departures moved the output toward the reference report rather than introducing an unsupported reference-level finding. In case A014, the rewrite omitted a predicted prosthetic crown at tooth 24 and redundantly repeated endodontic treatment at tooth 36; the reference described a coronal restoration rather than a crown at tooth 24. In case A067, the rewrite changed a predicted coronal restoration at tooth 43 to a prosthetic crown, matching the reference. These observations indicate that rewriting is not strictly lossless and can occasionally regularize imperfect predicted evidence, consistent with its modest mean F1 improvement. The retained deterministic template provides an auditable fallback and exposes such changes directly.

\section{Controlled Direct Generation Suite}
\subsection{Controlled interfaces and supervision}
We instantiate the direct generators with general-purpose Qwen3-8B \citep{yang2025qwen3} and clinically aligned Llama3-Med42-8B \citep{christophe2024med42}. Both use LoRA rank 16, completion-only loss for 12 epochs, deterministic decoding, and the training and evaluation split defined in the main paper. Two targets are used throughout: \emph{raw} denotes the observed report, and \emph{canonical} denotes the report obtained from reference-derived evidence through deterministic rendering and the frozen rewriter.

Direct 1 projects 72 frozen Stage~1 structure tokens into the language model. Direct 2 recomputes those tokens from the fixed-size CBCT input and updates encoder levels 2-4. Direct 3 uses the same global, relation, and tooth views and branch architectures as Stage~2 and initializes them from the corresponding primary Stage~2 checkpoints. During direct SFT, these visual modules and their feature adapter remain trainable, and their continuous output is supplied directly to the language decoder without a discrete evidence interface.

Direct 4 retains the complete Direct 3 pathway and adds evidence prediction heads. Its training objective is
\begin{equation}
\mathcal{L}_{\mathrm{direct+aux}}
=\mathcal{L}_{\mathrm{report}}
+\frac{\lambda}{3}
\left(\mathcal{L}_{\mathrm{global}}+
\mathcal{L}_{\mathrm{relation}}+
\mathcal{L}_{\mathrm{tooth}}\right),
\end{equation}
with \(\lambda=1\). Global and relation terms are masked BCE losses; the tooth term uses the same PU-weighted reliability values as the Stage~2 tooth labels. Ground-truth evidence is used only to form this training loss and is not supplied at inference.

Direct 5 retains the Direct 3 continuous-feature interface but freezes all Stage~2 branch parameters at their evidence-trained checkpoints. Only the feature adapter and language-model LoRA parameters are optimized. This control separates the effect of preserving the Stage~2 representation from the effect of exposing a thresholded, discrete evidence record to the report generator.

\subsection{Run-level results}
The main paper reports logical precision, recall, F1, BLEU-4, and METEOR for the complete comparison. Table~\ref{tab:direct-runs} provides logical-F1 for each model run. Canonical targets improve all ten decoder-interface pairs. Med42 improves the strongest direct result from \(.331\) with Qwen3 to \(.371\). Direct 5 reaches \(.327\) with Qwen3 and \(.360\) with Med42, and does not improve on trainable Direct 3 with either decoder.

\begin{table*}[t]
\centering
\small
\begin{tabular}{@{}llllccc@{}}
\toprule
Family & Visual interface & Auxiliary evidence & Target & Run 1 & Run 2 & Run 3 \\
\midrule
Direct Qwen3-8B & Direct 1: frozen Stage~1 & No & Raw & .235 & .242 & .209 \\
 & Direct 1: frozen Stage~1 & No & Canonical & .325 & .258 & .297 \\
 & Direct 2: fine-tuned Stage~1 & No & Raw & .244 & .265 & .251 \\
 & Direct 2: fine-tuned Stage~1 & No & Canonical & .283 & .355 & .327 \\
 & Direct 3: continuous Stage~2 features & No & Raw & .240 & .258 & .210 \\
 & Direct 3: continuous Stage~2 features & No & Canonical & .328 & .330 & .337 \\
 & Direct 4: Direct 3 + auxiliary evidence & Yes & Raw & .239 & .214 & .281 \\
 & Direct 4: Direct 3 + auxiliary evidence & Yes & Canonical & .341 & .315 & .338 \\
 & Direct 5: frozen Stage~2 encoder & No & Raw & .240 & .239 & .223 \\
 & Direct 5: frozen Stage~2 encoder & No & Canonical & .326 & .325 & .330 \\
\midrule
Direct Med42-8B & Direct 1: frozen Stage~1 & No & Raw & .249 & .240 & .241 \\
 & Direct 1: frozen Stage~1 & No & Canonical & .341 & .358 & .334 \\
 & Direct 2: fine-tuned Stage~1 & No & Raw & .275 & .304 & .256 \\
 & Direct 2: fine-tuned Stage~1 & No & Canonical & .375 & .351 & .386 \\
 & Direct 3: continuous Stage~2 features & No & Raw & .283 & .260 & .296 \\
 & Direct 3: continuous Stage~2 features & No & Canonical & .358 & .360 & .370 \\
 & Direct 4: Direct 3 + auxiliary evidence & Yes & Raw & .270 & .271 & .274 \\
 & Direct 4: Direct 3 + auxiliary evidence & Yes & Canonical & .358 & .364 & .372 \\
 & Direct 5: frozen Stage~2 encoder & No & Raw & .266 & .269 & .251 \\
 & Direct 5: frozen Stage~2 encoder & No & Canonical & .363 & .345 & .371 \\
\midrule
EviDent-CBCT & Consistency-projected evidence record & -- & Template & .401 & .393 & .401 \\
 & Consistency-projected evidence record & -- & Local rewrite & .405 & .399 & .403 \\
\bottomrule
\end{tabular}%
\caption{Run-level RadFact-Lite-Dental logical-F1 for the controlled report-generation comparison.}
\label{tab:direct-runs}
\end{table*}

\subsection{Case-level paired bootstrap}
We compare the final consistency-projected, locally rewritten EviDent-CBCT reports with the strongest direct configuration by unrounded mean logical-F1, Med42 Direct 2 with canonical supervision. For each system, case-level RadFact-Lite-Dental logical precision and recall are first averaged across its three model runs. Using random seed 2027, we jointly resample the matched case indices with replacement 100,000 times. Within every replicate, corpus-level precision and recall are averaged over the sampled cases and logical-F1 is recomputed as their harmonic mean. The percentile 95\% interval is taken from the resulting paired differences, \(F1_{\mathrm{EviDent}}-F1_{\mathrm{Direct}}\).

Using the final consistency-projected outputs and unrounded case-averaged values, the direct and EviDent-CBCT aggregate logical-F1 scores are .3706 and .4023, respectively. Their paired difference is .0317, with a 95\% bootstrap confidence interval of [.0047, .0586]. This interval quantifies case-sampling uncertainty after averaging model runs; it is not a confidence interval over training seeds.

The comparison between Direct 3 and Direct 4 separates auxiliary evidence supervision from an evidence bottleneck. Adding the auxiliary objective changes Qwen3 logical-F1 from \(.236\) to \(.244\) for raw targets and from \(.331\) to \(.331\) for canonical targets; the corresponding Med42 changes are \(.280\) to \(.271\) and \(.362\) to \(.365\). Freezing the Stage~2 encoders in Direct 5 yields \(.234\) and \(.327\) with Qwen3, and \(.262\) and \(.360\) with Med42. Neither auxiliary supervision nor preservation of continuous Stage~2 features reproduces the .402 obtained when the consistency-projected discrete record is the required report input.

\subsection{Sensitivity to multiple reports per volume}
The fixed-reference evaluation selects the report with the lexicographically smallest identifier for each volume. As an additional robustness analysis, we evaluate each candidate against every report associated with its volume; 39 of the 63 validation volumes have two reports and 24 have one (102 reports in total). If \(P_{ij}\) and \(R_{ij}\) are the logical precision and recall for candidate \(i\) against its \(j\)-th reference, we first compute
\begin{equation}
\bar P_i=\frac{1}{m_i}\sum_{j=1}^{m_i}P_{ij},\qquad
\bar R_i=\frac{1}{m_i}\sum_{j=1}^{m_i}R_{ij},
\end{equation}
then average \(\bar P_i\) and \(\bar R_i\) over the 63 volumes and compute their harmonic mean. Thus, every volume retains equal weight regardless of its number of reports.

\begin{table*}[t]
\centering
\small
\begin{tabular}{@{}llccc@{}}
\toprule
Reference protocol & Candidate output & Logical P & Logical R & Logical F1 \\
\midrule
Fixed lexicographic & Med42 Direct 2, canonical (direct best) & \(.388\pm.019\) & \(.354\pm.017\) & \(.371\pm.018\) \\
 & EviDent-CBCT template & \(.372\pm.002\) & \(.428\pm.008\) & \(.398\pm.004\) \\
 & EviDent-CBCT local rewrite & \(.384\pm.002\) & \(.423\pm.004\) & \(.402\pm.003\) \\
\midrule
Mean over all references & Med42 Direct 2, canonical (direct best) & \(.361\pm.013\) & \(.310\pm.013\) & \(.333\pm.013\) \\
 & EviDent-CBCT template & \(.369\pm.007\) & \(.379\pm.012\) & \(.374\pm.009\) \\
 & EviDent-CBCT local rewrite & \(.374\pm.003\) & \(.373\pm.011\) & \(.374\pm.006\) \\
\bottomrule
\end{tabular}
\caption{Sensitivity of the final comparison to the reference protocol, reported as mean\(\pm\)std over three model runs. Under mean-over-references evaluation, logical precision and recall are averaged within each volume before corpus aggregation.}
\label{tab:multi-reference-sensitivity}
\end{table*}

Under mean-over-references evaluation, both EviDent-CBCT realizations round to .374 logical-F1, compared with .333 for the strongest direct generator. For the final locally rewritten output, the unrounded paired advantage over direct generation is .0402, with a 95\% bootstrap confidence interval of [.0170, .0637] using the same 100,000 volume-level resamples. The rewrite--template difference is negligible (\(-.0004\), 95\% CI [\(-.0074\), .0065]), indicating that the advantage over direct generation persists independently of the local realizer and reference selection.

\section{Global and Relation Fusion Sweep}
The main paper fixes the global branch to \emph{scratch} and the relation branch to \emph{hybrid} (\texttt{g\_scratch\_r\_hybrid}). All rows share the tooth-base branch. Table~\ref{tab:fusion} reports a single-run architecture sweep using the LLM-free merged Stage~2 logical set metric.

\begin{table}[t]
\centering
\small
\begin{tabular}{@{}lcccc@{}}
\toprule
Global & Relation & Logical P & Logical R & Logical F1 \\
\midrule
hybrid & hybrid & .653 & .692 & .672 \\
hybrid & scratch & .651 & .691 & .670 \\
hybrid & visual & .651 & .695 & .672 \\
scratch & hybrid & .652 & .696 & .673 \\
scratch & visual & .650 & .699 & .674 \\
visual & hybrid & .655 & .702 & .678 \\
\bottomrule
\end{tabular}
\caption{Global/relation interface sweep using merged Stage~2 exact-tuple agreement. The narrow range does not support a claim that one fusion choice is intrinsically superior.}
\label{tab:fusion}
\end{table}

The observed logical-F1 range is 0.670-0.678. We retain \texttt{g\_scratch\_r\_hybrid} as the predefined base for the tooth ablations and do not treat this narrow sweep as evidence for a general fusion mechanism.

\section{Evaluation Semantics}
\subsection{Merged Stage~2 logical set-F1}
For case $k$, let $C_k$ be the set of predicted positive tuples and $R_k$ the positive tuples derived from its reference report, pooled across tooth, global, and relation branches within the evaluation mask. Exact structured matching gives
\begin{equation}
P_k=\frac{|C_k\cap R_k|}{|C_k|},\qquad
R_k^{\mathrm{rec}}=\frac{|C_k\cap R_k|}{|R_k|}.
\end{equation}
Zero denominators contribute zero, matching the RadFact-Lite aggregation convention. The reported logical precision and recall are the means of $P_k$ and $R_k^{\mathrm{rec}}$ over cases; logical set-F1 is the harmonic mean of those two corpus-level quantities. This is an LLM-free structured-agreement metric. It is not branch macro-F1 and is not the report-level RadFact score.

\subsection{RadFact-Lite-Dental}
RadFact-Lite-Dental is a dentist-guided evaluation extension built from RadFact-Lite v0.1.0~\citep{radfactlite010}, a lightweight implementation of the text-only RadFact procedure~\citep{bannur2024maira2}. It retains RadFact's bidirectional proposition-entailment protocol and corpus aggregation while adapting proposition construction and entailment assessment to dento-maxillofacial reports. GPT-5.6-Luna first parses each report into complete, image-verifiable dental propositions. The parser atomizes tooth lists but preserves qualifiers such as laterality, severity, extent, uncertainty, visibility, treatment status, and anatomical relations; bridge spans remain relational propositions rather than independent tooth labels.

Before LLM assessment, conservative normalization maps equivalent dotted and undotted FDI notation to a common form, removes semantic duplicates, and resolves exact or closed-form positive entailments. The deterministic layer never rejects an unresolved comparison. Remaining hypotheses are assessed jointly for each case and direction with dental-specific instructions that require agreement in tooth position, anatomy, side, finding type, polarity, and clinically meaningful qualifiers. Parsing and entailment outputs are cached for identical normalized inputs, and this final RadFact-Lite-Dental configuration is used for all report-level evaluations.

For report-level evaluation, the report with the lexicographically smallest identifier was fixed as the reference for each volume and used across all systems and evaluator runs.

Official ODIN hidden-test RadFact uses a different evaluation protocol and is therefore reported separately without direct numerical comparison to RadFact-Lite-Dental.

\paragraph{Author audit of RadFact-Lite-Dental.}
We conducted an author audit of 30 candidate-reference report pairs, sampling ten cases from each tertile of total proposition count. For each pair, the complete reports, proposition boundaries, and every bidirectional entailment decision were reviewed to establish author-audited gold scores. Table~\ref{tab:radfact-dental-audit} compares these scores with the original RadFact-Lite v0.1.0 implementation using GPT-4o without dental FDI normalization and with the final RadFact-Lite-Dental evaluator. Under the same corpus-level aggregation, RadFact-Lite-Dental differs from the author-audited logical-F1 by \(-.009\), with a paired case-bootstrap 95\% confidence interval of \([-.030,.012]\).

\begin{table*}[t]
\centering
\small
\begin{tabular}{@{}lccccc@{}}
\toprule
Evaluation & Logical P & Logical R & Logical F1 & F1 MAE & Pearson \(r\) \\
\midrule
Author-audited gold & .364 & .468 & .410 & -- & -- \\
RadFact-Lite v0.1.0 & .371 & .458 & .410 & .091 & .587 \\
RadFact-Lite-Dental & .366 & .444 & .401 & .043 & .881 \\
\bottomrule
\end{tabular}
\caption{Agreement with author-audited gold scores on 30 stratified report pairs. MAE and Pearson correlation are computed over case-level logical-F1.}
\label{tab:radfact-dental-audit}
\end{table*}

The original evaluator nearly reproduces the gold cohort-level F1 (.410 versus .410), but this agreement does not extend to individual cases: its case-level MAE is .091, Pearson \(r=.587\), and maximum absolute error is .307. Its 19 overestimates and 11 underestimates largely cancel in the aggregate. The final dental evaluator reduces MAE by 52.7\% to .043, increases Pearson correlation to .881, and reduces the maximum absolute error to .140. Its Spearman correlation is \(\rho=.861\), and 20 of 30 cases fall within .05 F1 and 26 within .10, compared with 12 and 19 cases for the original evaluator. Across 1,375 fixed-proposition entailment decisions, the final evaluator reaches .949 accuracy, .941 balanced accuracy, .931 entailment F1, and Cohen's \(\kappa=.891\) against the author audit. Thus, its small aggregate offset accompanies substantially closer case-level agreement and more faithful case ordering.

\begin{table*}[t]
\centering
\scriptsize
\setlength{\tabcolsep}{1.5mm}
\begin{tabular}{@{}lccccccccc@{}}
\toprule
& \multicolumn{3}{c}{\texttt{pu-weak} (base)} & \multicolumn{3}{c}{\texttt{explicit}} & \multicolumn{3}{c}{\texttt{hard}} \\
\cmidrule(lr){2-4} \cmidrule(lr){5-7} \cmidrule(lr){8-10}
Finding & Logical P & Logical R & Logical F1 & Logical P & Logical R & Logical F1 & Logical P & Logical R & Logical F1 \\
\midrule
Missing tooth & \(.412\pm.021\) & \(.622\pm.087\) & \(\mathbf{.494\pm.025}\) & \(.346\pm.027\) & \(.524\pm.075\) & \(.413\pm.016\) & \(.368\pm.024\) & \(.694\pm.084\) & \(.478\pm.002\) \\
Implant & \(.325\pm.013\) & \(.617\pm.021\) & \(.426\pm.014\) & \(.432\pm.092\) & \(.543\pm.057\) & \(\mathbf{.473\pm.046}\) & \(.338\pm.037\) & \(.580\pm.057\) & \(.426\pm.035\) \\
Crown & \(.452\pm.013\) & \(.646\pm.065\) & \(\mathbf{.530\pm.012}\) & \(.233\pm.101\) & \(.646\pm.053\) & \(.331\pm.111\) & \(.427\pm.031\) & \(.663\pm.079\) & \(.516\pm.005\) \\
Restoration & \(.360\pm.018\) & \(.665\pm.026\) & \(.466\pm.011\) & \(.161\pm.002\) & \(.943\pm.010\) & \(.275\pm.004\) & \(.424\pm.030\) & \(.542\pm.048\) & \(\mathbf{.474\pm.016}\) \\
Endodontic treatment & \(.413\pm.036\) & \(.595\pm.097\) & \(\mathbf{.482\pm.010}\) & \(.385\pm.083\) & \(.356\pm.035\) & \(.367\pm.049\) & \(.382\pm.042\) & \(.595\pm.043\) & \(.464\pm.035\) \\
Prosthesis-related & \(.604\pm.024\) & \(.789\pm.029\) & \(\mathbf{.683\pm.005}\) & \(.611\pm.167\) & \(.513\pm.042\) & \(.549\pm.065\) & \(.649\pm.042\) & \(.693\pm.054\) & \(.668\pm.007\) \\
Impacted/disodontiasis & \(.750\pm.035\) & \(.920\pm.040\) & \(\mathbf{.826\pm.037}\) & \(.685\pm.071\) & \(.853\pm.023\) & \(.759\pm.052\) & \(.756\pm.046\) & \(.893\pm.046\) & \(.817\pm.009\) \\
Caries & \(.149\pm.160\) & \(.467\pm.321\) & \(\mathbf{.121\pm.029}\) & \(.014\pm.001\) & \(.667\pm.153\) & \(.027\pm.001\) & \(.051\pm.019\) & \(.433\pm.115\) & \(.089\pm.026\) \\
Root remnant & \(.095\pm.064\) & \(.111\pm.048\) & \(.090\pm.028\) & \(.104\pm.032\) & \(.361\pm.096\) & \(\mathbf{.159\pm.043}\) & \(.045\pm.016\) & \(.306\pm.048\) & \(.078\pm.025\) \\
Root fracture & \(.002\pm.002\) & \(.667\pm.577\) & \(.004\pm.004\) & \(.004\pm.001\) & \(1.000\pm.000\) & \(\mathbf{.009\pm.003}\) & \(.001\pm.002\) & \(.333\pm.577\) & \(.002\pm.004\) \\
Lesion-related & \(.105\pm.017\) & \(.444\pm.030\) & \(.170\pm.024\) & \(.045\pm.014\) & \(.427\pm.339\) & \(.074\pm.009\) & \(.127\pm.024\) & \(.325\pm.039\) & \(\mathbf{.180\pm.020}\) \\
\midrule
\textbf{Branch logical set} & \(.434\pm.028\) & \(.538\pm.022\) & \(\mathbf{.480\pm.007}\) & \(.405\pm.057\) & \(.328\pm.035\) & \(.359\pm.018\) & \(.408\pm.017\) & \(.548\pm.030\) & \(.467\pm.017\) \\
\bottomrule
\end{tabular}%
\caption{Class-wise tooth-branch Stage~2 performance under the three non-mention policies, reported as mean$\pm$std over three model runs. The final row is the case-grouped branch logical set score over all tooth categories.}
\label{tab:tooth-classwise-stage2}
\end{table*}

\begin{table*}[t]
\centering
\scriptsize
\begin{tabular}{@{}llrrrrr@{}}
\toprule
Branch & Finding & Known & Positive & Logical P & Logical R & Logical F1 \\
\midrule
Global & Partial maxilla coverage & 54 & 27 & .579 & .815 & .677 \\
 & Maxilla excluded & 57 & 21 & .654 & .810 & .723 \\
 & Maxillary sinuses excluded & 30 & 14 & .933 & 1.000 & .966 \\
 & Mandibular condyles excluded & 59 & 58 & .983 & 1.000 & .991 \\
 & Periodontitis & 19 & 19 & 1.000 & 1.000 & 1.000 \\
 & Maxillary atrophy & 8 & 6 & 1.000 & .833 & .909 \\
 & Mandibular atrophy & 39 & 34 & .944 & 1.000 & .971 \\
 & Severe atrophy & 32 & 12 & .750 & .750 & .750 \\
 & Osteolytic lesion & 33 & 17 & .778 & .824 & .800 \\
 & Osteosclerotic lesion & 23 & 2 & .286 & 1.000 & .444 \\
 & Sinus pathology & 16 & 6 & .750 & 1.000 & .857 \\
 & Poor quality/artifact & 2 & 2 & 1.000 & 1.000 & 1.000 \\
\cmidrule(lr){2-7}
 & \textbf{Global logical set} & -- & -- & .890 & .878 & \textbf{.884} \\
\midrule
Relation & Tooth 38 close to left IAC & 19 & 18 & .947 & 1.000 & .973 \\
 & Tooth 48 close to right IAC & 13 & 13 & 1.000 & 1.000 & 1.000 \\
 & Any tooth close to IAC & 20 & 20 & 1.000 & 1.000 & 1.000 \\
 & Lesion close to IAC & 1 & 1 & 1.000 & 1.000 & 1.000 \\
 & Left IAC predominantly lingual & 29 & 22 & .786 & 1.000 & .880 \\
 & Left IAC predominantly buccal & 27 & 6 & .263 & .833 & .400 \\
 & Right IAC predominantly lingual & 26 & 19 & .760 & 1.000 & .864 \\
 & Right IAC predominantly buccal & 25 & 5 & .364 & .800 & .500 \\
\cmidrule(lr){2-7}
 & \textbf{Relation logical set} & -- & -- & .512 & .563 & \textbf{.537} \\
\bottomrule
\end{tabular}
\caption{Class-wise Stage~2 performance of the base global and relation branches. Counts expose the support available for each category; branch logical-set rows are case-grouped and are not averages of the class rows.}
\label{tab:base-other-classwise-stage2}
\end{table*}

\section{Class-wise Stage~2 Branch Analysis}
The three supervision policies alter only the tooth branch; the crop geometry, gated hybrid encoder, metal-sensitive channel, global branch, and relation branch are otherwise fixed. We therefore compare \texttt{pu-weak}, \texttt{explicit}, and \texttt{hard} directly at the tooth interface. For a tooth category, a tuple is defined by its FDI position and finding label. Table~\ref{tab:tooth-classwise-stage2} reports exact-tuple precision, recall, and F1 on the strict evaluation mask, with unmentioned eligible tooth attributes treated as negatives.

The branch-level ordering matches the aggregate tooth-branch ablation in the main paper: \texttt{pu-weak} gives the highest tooth logical set-F1, followed by \texttt{hard} and then \texttt{explicit}. Its advantage is not uniform across every rare label, but it obtains the highest class F1 for six of the 11 categories, including missing teeth, crowns, endodontic treatment, prosthesis-related findings, impaction/disodontiasis, and caries. The \texttt{explicit} model often retains recall while losing precision under the strict mask, most visibly for restorations (P=.161, R=.943) and caries (P=.014, R=.667), showing that ignored non-mentions leave many unsupported positives unconstrained. Full-weight negatives recover much of this precision, while reduced-reliability negatives provide a better aggregate balance. Root fracture, caries, and root remnant have only 1, 10, and 12 positive validation instances, respectively; their policy-specific ordering should therefore be interpreted as descriptive rather than as a stable ranking.

Table~\ref{tab:base-other-classwise-stage2} completes the branch view for the fixed base system. \emph{Known} is the number of eligible supervised entries and \emph{Positive} is the number of reference-derived positives within those entries.

The global branch is strongest for common coverage and atrophy statements, reaching .884 branch logical set-F1. Osteosclerotic lesions remain precision-limited, while the perfect scores for periodontitis and poor-quality scans are supported by no observed negatives and only 19 and two positives, respectively. In the relation branch, tooth-IAC proximity and lingual-course labels are recovered reliably on their eligible entries, whereas the less frequent buccal-course categories remain precision-limited. The lower relation logical set-F1 (.537) reflects case-level agreement over the complete predicted relation set and is therefore more demanding than the high pooled F1 of several individual, positive-heavy categories.

\section{Qualitative Report Comparisons}
\label{sec:qualitative_examples}

\begin{figure*}[t]
    \centering
    \includegraphics[
        width=\textwidth,
        height=0.82\textheight,
        keepaspectratio
    ]{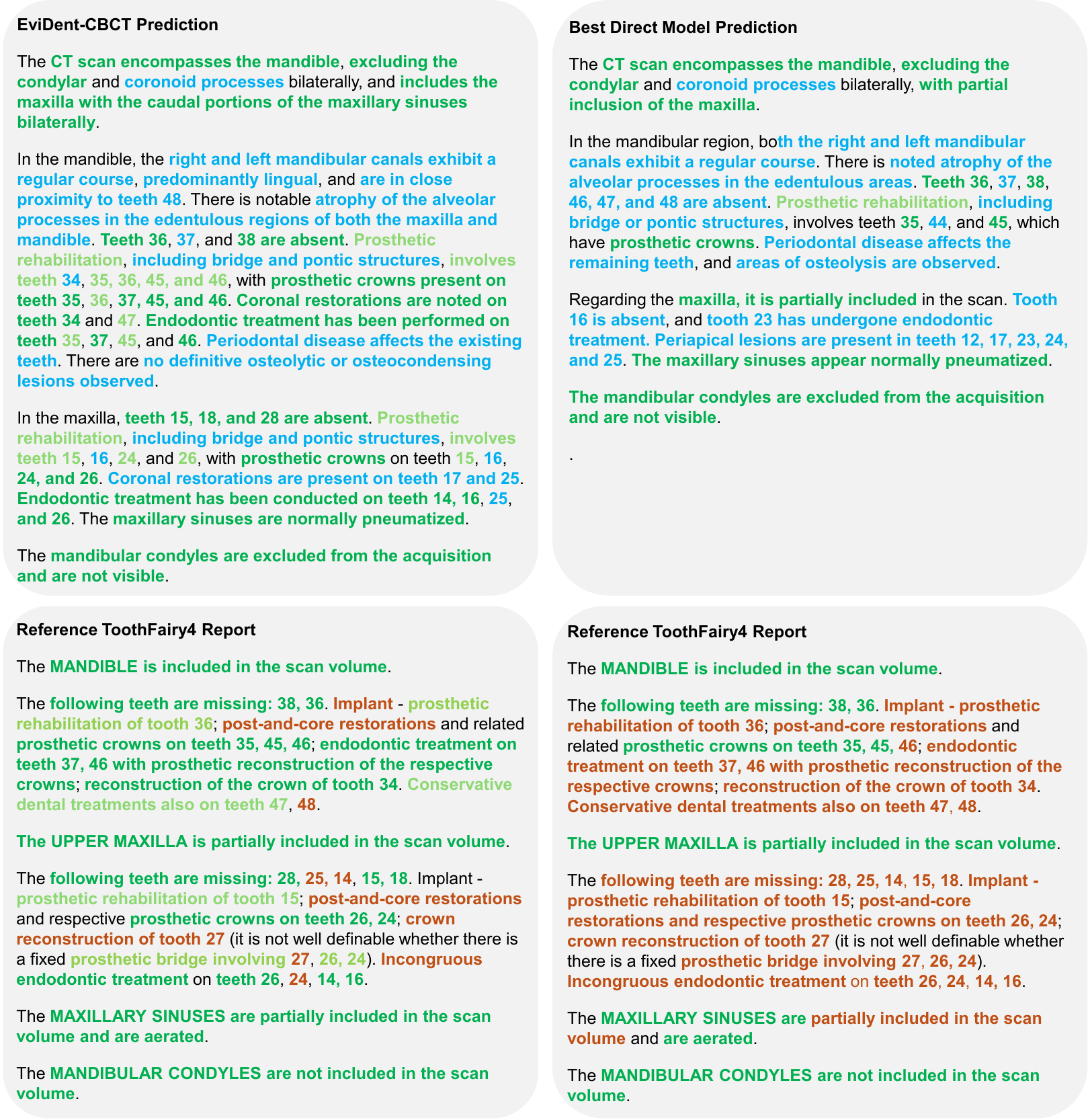}
    \caption{
    A treatment-rich case with partial maxillary coverage. EviDent-CBCT recovers broader reference-supported tooth-status, restorative, and endodontic findings than the best direct model, although both predictions contain unsupported or mismatched details. In each composite, predictions appear above their annotated references. Dark green indicates a match, light green a partial or underspecified match, blue an unsupported or contradictory prediction, and red an unrecovered reference fact.
    }
    \label{fig:qualitative_demo1}
\end{figure*}

\begin{figure*}[t]
    \centering
    \includegraphics[
        width=\textwidth,
        height=0.82\textheight,
        keepaspectratio
    ]{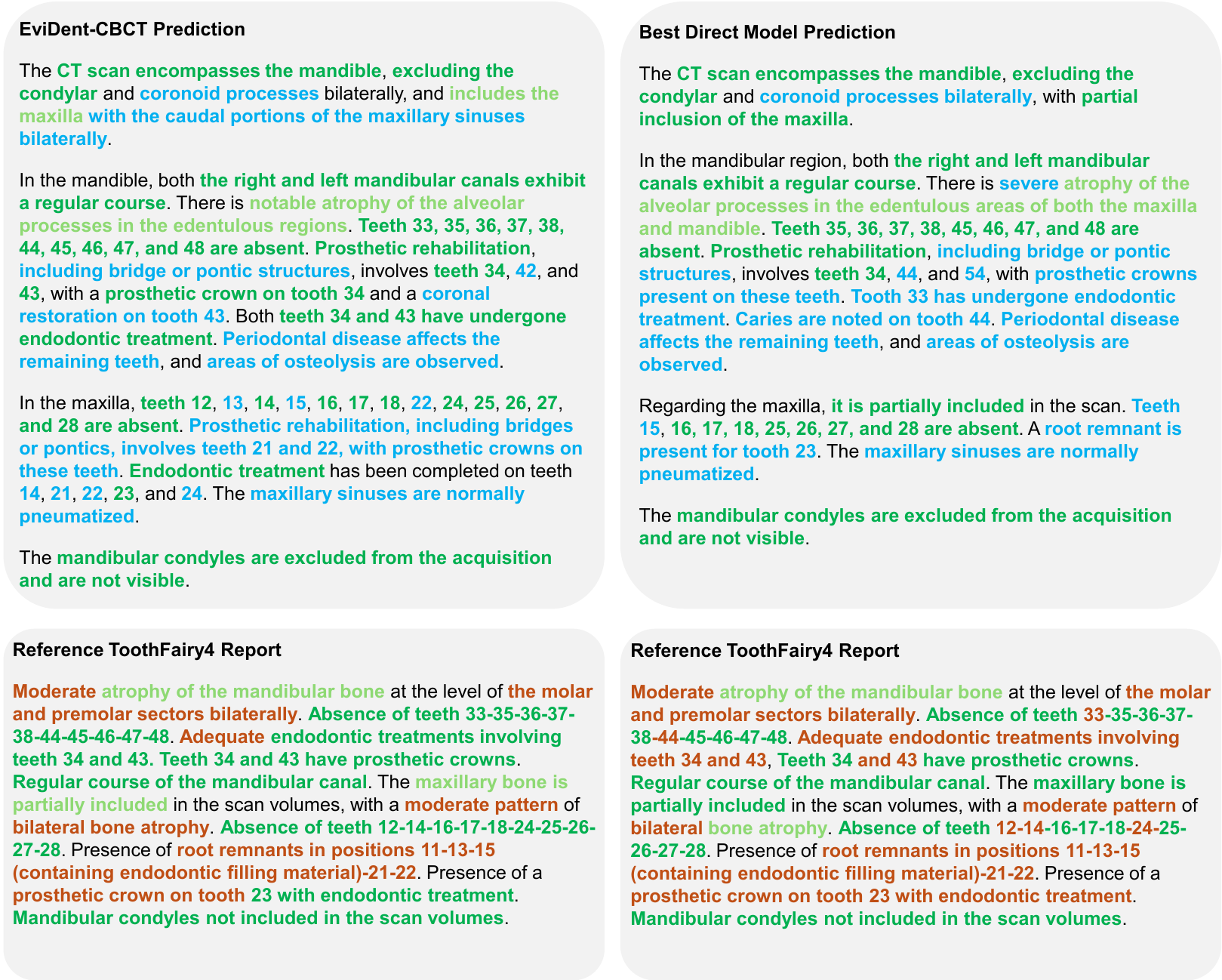}
    \caption{
    A heavily edentulous case with a complex treatment history. EviDent-CBCT recovers the complete set of mandibular missing teeth and correctly localizes several restorative and endodontic findings. The direct model omits or misclassifies more tooth-level findings, while both models remain challenged by maxillary root remnants. In each composite, predictions appear above their annotated references. Dark green indicates a match, light green a partial or underspecified match, blue an unsupported or contradictory prediction, and red an unrecovered reference fact.
    }
    \label{fig:qualitative_demo2}
\end{figure*}

\begin{figure*}[t]
    \centering
    \includegraphics[
        width=\textwidth,
        height=0.82\textheight,
        keepaspectratio
    ]{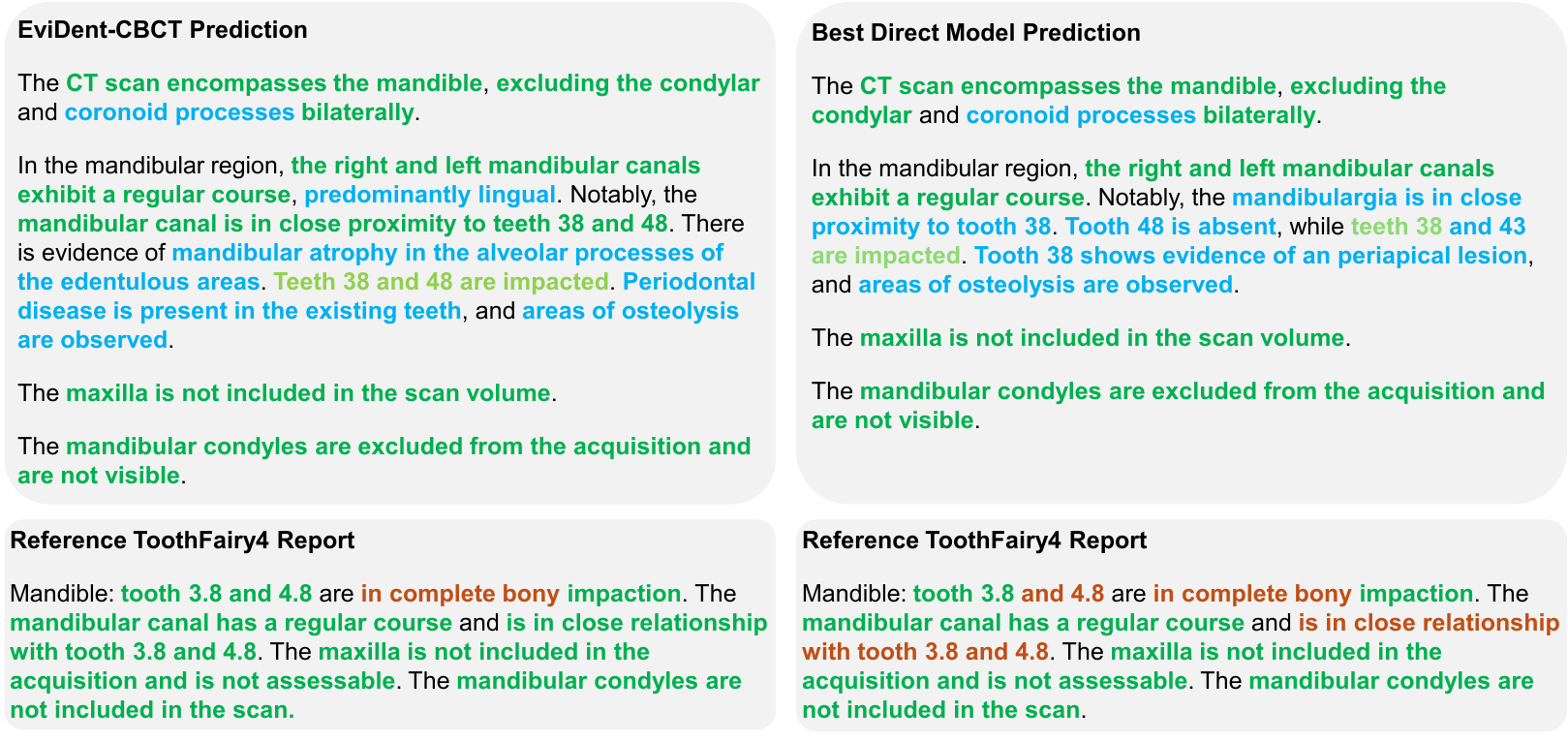}
    \caption{
    A mandibular-only case with bilateral complete bony impaction of teeth 38 and 48. EviDent-CBCT captures both impacted teeth and their relationships with the mandibular canal, but omits the complete-bony qualifier and introduces several unsupported findings. The direct model misclassifies tooth 48 as absent and adds unsupported tooth-level abnormalities. In each composite, predictions appear above their annotated references. Dark green indicates a match, light green a partial or underspecified match, blue an unsupported or contradictory prediction, and red an unrecovered reference fact.
    }
    \label{fig:qualitative_demo3}
\end{figure*}

\begin{figure*}[t]
    \centering
    \includegraphics[
        width=\textwidth,
        height=0.82\textheight,
        keepaspectratio
    ]{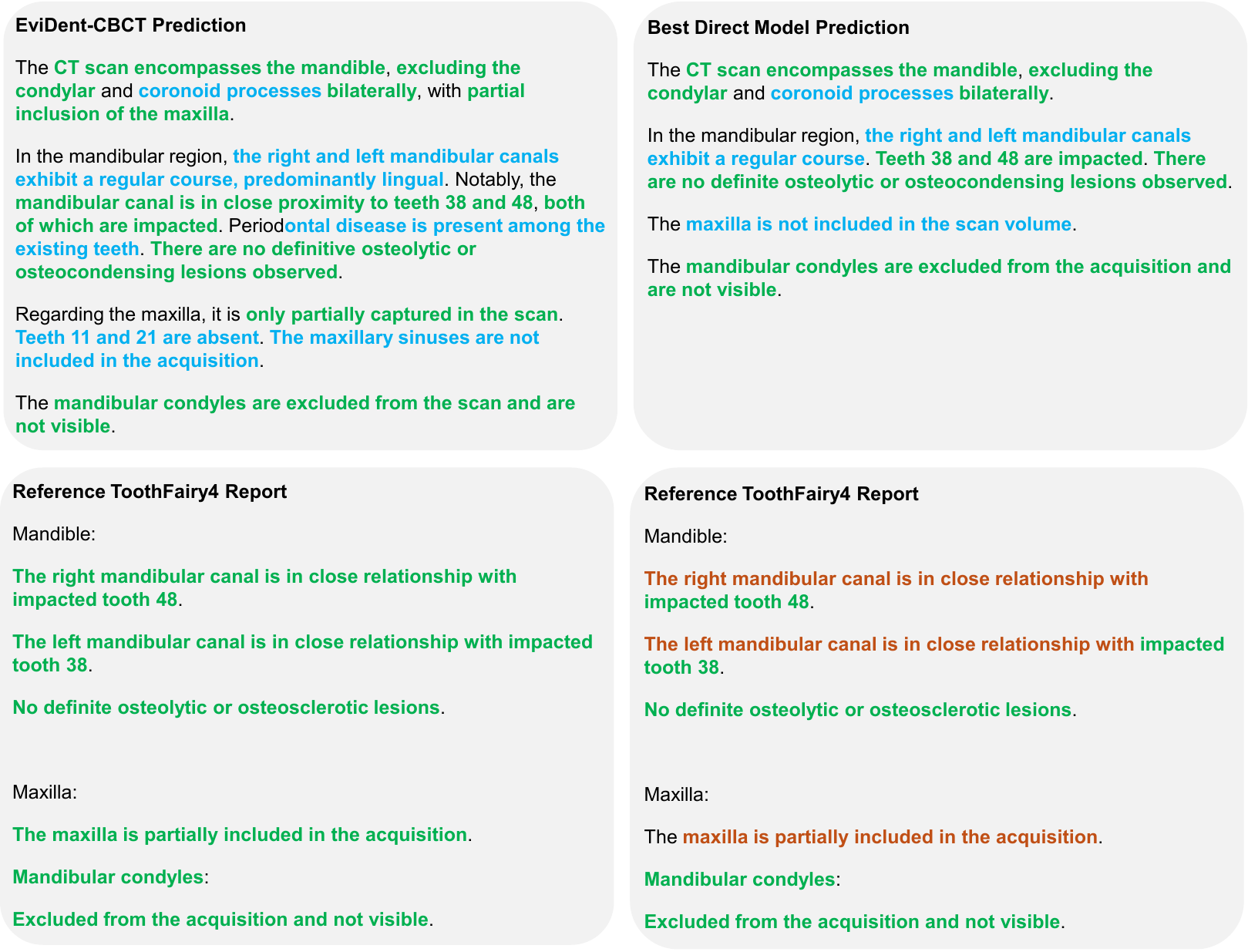}
    \caption{
    A case with bilateral canal--impacted-third-molar relationships and partial maxillary inclusion. EviDent-CBCT covers all core reference propositions, including both canal relationships, the negative osseous finding, and the acquisition coverage. The direct model identifies the impactions but omits their canal relationships and incorrectly reports that the maxilla is not included. In each composite, predictions appear above their annotated references. Dark green indicates a match, light green a partial or underspecified match, blue an unsupported or contradictory prediction, and red an unrecovered reference fact.
    }
    \label{fig:qualitative_demo4}
\end{figure*}

Figures~\ref{fig:qualitative_demo1}--\ref{fig:qualitative_demo4} compare representative EviDent-CBCT predictions with those of the best direct model. In each composite, the upper-left and upper-right panels show the respective predictions, while the lower panels repeat the reference report with annotations relative to the prediction above. Dark green denotes a supported match, light green a partially correct or underspecified statement, blue an unsupported or contradictory prediction, and red a reference fact not recovered by the corresponding prediction. Uncolored text provides linguistic context. 

